\documentclass[twoside,11pt]{article}

\usepackage{jmlr2e}

\usepackage{amsmath}
\usepackage[dvipsnames]{xcolor}
\usepackage[linesnumbered,ruled,vlined]{algorithm2e}
\usepackage[noend]{algpseudocode}
\usepackage{multicol,multirow}
\usepackage{caption}
\usepackage{subcaption}
\usepackage{paracol}
\usepackage{graphicx}
\usepackage{arydshln}
\usepackage{hhline}
\usepackage[title]{appendix}
\usepackage{enumitem}
\usepackage{pdfpages}
\usepackage{float}
\usepackage{hyperref}
\usepackage[title]{appendix}
\usepackage{titletoc}

\hypersetup{
	colorlinks=true,
	linkcolor=blue,
	filecolor=magenta,      
	urlcolor=cyan,
	pdfpagemode=FullScreen,
	hidelinks
}

\DeclareMathOperator{\Cov}{Cov}
\DeclareMathOperator{\E}{E}
\DeclareMathOperator{\Var}{Var}

\DeclareMathOperator{\Lap}{Lap}

\newcommand{\mbH}{\mathbb{H}}
\newcommand{\mbR}{\mathbb{R}}

\newcommand{\mcC}{\mathcal{C}}
\newcommand{\mcD}{\mathcal{D}}
\newcommand{\mcH}{\mathcal{H}}

\newcommand \mcT{\mathcal{T}}
\newcommand{\mcX}{\mathcal{X}}

\usepackage{lastpage}
\jmlrheading{25}{2024}{1-\pageref{LastPage}}{12/22; Revised
7/24}{10/24}{22-1384}{Haotian Lin and Matthew Reimherr}
\ShortHeadings{Pure Differential Privacy for Functional Summaries}{Lin and Reimherr}

\begin{document}

\title{Pure Differential Privacy for Functional Summaries with a Laplace-like Process}

\author{\name Haotian Lin \email hzl435@psu.edu \\
        \addr Department of Statistics, \\
       The Pennsylvania State University, \\
       University Park, PA 16802, USA 
       \vspace{0.2cm}
       \\
        \name Matthew Reimherr \email mreimherr@psu.edu \\
       \addr Department of Statistics, \\
       The Pennsylvania State University, \\
       University Park, PA 16802, USA 
       \vspace{0.2cm}}

\editor{Jie Peng}

\maketitle

\begin{abstract}
Many existing mechanisms for achieving differential privacy (DP) on infinite-dimensional functional summaries typically involve embedding these functional summaries into finite-dimensional subspaces and applying traditional multivariate DP techniques. These mechanisms generally treat each dimension uniformly and struggle with complex, structured summaries. This work introduces a novel mechanism to achieve pure DP for functional summaries in a separable infinite-dimensional Hilbert space, named the \textit{Independent Component Laplace Process} (ICLP) mechanism. This mechanism treats the summaries of interest as truly infinite-dimensional functional objects, thereby addressing several limitations of the existing mechanisms. Several statistical estimation problems are considered, and we demonstrate how one can enhance the utility of private summaries by oversmoothing the non-private counterparts. Numerical experiments on synthetic and real datasets demonstrate the effectiveness of the proposed mechanism.
\end{abstract}

\begin{keywords}
Differential Privacy, Functional Data Analysis, Hilbert Space, Reproducing Kernel Hilbert Space, Infinite-Dimensional Statistics.
\end{keywords}

\section{Introduction}

Data privacy has garnered critical attention in the last decade as substantial amounts of individualized data are collected. The most widely used paradigm in formal data privacy is \textit{differential privacy} (DP), introduced by \citet{dwork2006calibrating}. DP provides a rigorous and interpretable definition of data privacy, as it limits the amount of information attackers can infer from publicly released database queries. Numerous mechanisms have been developed for conventional data settings, such as scalar or vector-valued data. However, advances in technologies enable us to collect and process densely observed data over some temporal or spatial domains, which are coined \textit{functional data} to differentiate them from classic multivariate data \citep{ramsay2005functional,kokoszka2017introduction,ferraty2011oxford}. Although functional data analysis has been proven useful in various fields, such as economics, finance, and genetics, and has been widely researched in the statistical community, there are only a few works concerning privacy preservation within the realm of functional data.

When the statistical summaries are finite-dimensional, additive noise mechanisms are the most commonly used mechanisms to achieve DP, which privatize statistical summaries by adding calibrated noise from predetermined distributions, e.g., Laplace and Gaussian mechanisms \citep{dwork2006calibrating,dwork2014algorithmic}. 
In this paper, we are concerned with establishing an additive noise mechanism for functional summaries, namely infinite-dimensional summaries, to achieve $\epsilon$-DP. Given the challenge that functional summaries are typically infinite-dimensional, most existing mechanisms embed the non-private summaries into a finite-dimensional subspace by using finite basis expansions to approximate summaries and applying classical multivariate privacy tools, such as perturbing the expansion coefficients with i.i.d. noise \citep{zhang2012functional,wang2013efficient,chandrasekaran2014faster,alda2017bernstein}. This finite-dimensional embedding process is typically unavoidable, as when the summaries are infinite-dimensional, adding i.i.d. noise to each dimension is not even feasible if one wants the private summaries to remain in a specific infinite-dimensional function space.
However, these mechanisms have several weaknesses. First, determining the dimension of the subspace is crucial, as it plays a trade-off role between utility and privacy. While data-driven approaches might cause potential privacy leakage, a predetermined dimension will lack adaptation to the data, potentially failing to capture the structure or shape of the functional summaries or injecting excess noise. Second, in multivariate settings,
classical privacy tools that add i.i.d. noise to each dimension treat all dimensions equally and allocate the privacy budget over each dimension uniformly, failing to recognize the different levels of importance of coefficients across different dimensions and thus injecting excess noise for ``more important" dimensions. This substantially degrades the utility and robustness of private functional summaries. Some previous works have shown that capturing the covariance structure in the data might be able to reduce the amount of noise injected in specific scenarios \citep{hardt2010geometry,awan2021structure}.

\subsection{Our Contributions} 
To overcome the downsides inherent in DP mechanisms that rely on finite-dimensional embedding, we introduce a mechanism that treats both the functional summary and the privacy noise as truly infinite-dimensional functional objects. Concretely, our contributions can be summarized as follows:
\begin{enumerate}
    \item We propose an $\epsilon$-DP mechanism by perturbing functional summaries with a random element called the \textit{Independent Component Laplace Process} (ICLP) and name this mechanism \textit{the ICLP mechanism}. We establish the feasibility of the ICLP mechanism (meaning it can achieve DP) in an infinite-dimensional separable Hilbert space, $\mbH$, by characterizing a subspace of $\mbH$ and showing that the feasibility holds if and only if the difference between two functional summaries based on adjacent datasets resides in this subspace. We also show how the proposed mechanism applies to the space of continuous functions, even though this space is not a Hilbert space.

    \item We provide strategies based on regularized empirical risk minimization (regularized ERM) to obtain qualified functional summaries for the ICLP mechanism. We uncover the role regularization plays in the trade-off between utility and privacy. Specifically, we show that one can achieve $\epsilon$-DP with a matching order (or even a lower order) of privacy error as the estimation error by slightly oversmoothing the functional summaries in the functional mean protection problem. We also show that the application can go beyond classic functional data settings, as it is also applicable to the realm of more classic non-parametric smoothing problems like kernel density estimation. 
    
    \item To obtain privacy-safe regularization parameters in regularized ERM, we propose a privacy-safe selection approach so that choosing the parameters is only tied to the covariance structure of the ICLP noise, thus achieving end-to-end privacy. This approach overcomes the potential privacy leakage in conventional data-driven methods.
\end{enumerate}
The proposed mechanism differentiates itself from existing mechanisms, such as \citet{zhang2012functional,dwork2014algorithmic,alda2017bernstein}, that rely on finite-dimensional embedding and treat each dimension uniformly by adding i.i.d. noise to each dimension in the following senses. First, the ICLP mechanism avoids finite-dimensional subspace embeddings and frees the assumption that every dataset in the database shares the same finite-dimensional subspace. Second, unlike existing mechanisms, the ICLP mechanism treats each dimension heterogeneously, allowing it to achieve a more effective noise injection process while handling truly infinite-dimensional functional summaries and noise.

\subsection{Related Works}
In the overlap of functional summaries and differential privacy, the landmark paper is \citet{hall2013differential}, which provided a framework for achieving $(\epsilon,\delta)$-DP on infinite-dimensional functional objects but focused on a finite grid of evaluation points. The follow-up work in \citet{mirshani2019formal} pushed \citet{hall2013differential}'s result forward and established $(\epsilon,\delta)$-DP over the full functional path for objects in Banach spaces. In more general spaces, \citet{reimherr2019elliptical} considered elliptical perturbations to achieve $(\epsilon,\delta)$-DP in locally convex vector spaces, including all Hilbert spaces, Banach spaces, and Fr\'echet spaces. They also showed the impossibility of achieving $\epsilon$-DP for infinite-dimensional functional objects with elliptical distributions.

Turning to $\epsilon$-DP, a series of works has been proposed by resorting to finite-dimensional representations, such as polynomial bases, trigonometric bases, or Bernstein polynomial bases, to approximate target functional summaries \citep{wang2013efficient,chandrasekaran2014faster,alda2017bernstein} and loss functions \citep{zhang2012functional}. These mechanisms then perturb the expansion coefficients in $\mbR^{m}$ via the $m$-dimensional i.i.d. Laplace mechanism \citep{dwork2014algorithmic}. Privatizing $m$-dimensional coefficients is feasible through the $K$-norm mechanism \citep{hardt2010geometry,awan2021structure}, which encompasses the multivariate i.i.d. Laplace mechanism as a particular instance. However, to the best of our knowledge, no existing literature combines the $K$-norm mechanism with finite-dimensional embedding techniques to achieve DP for functional summaries. In addition to additive noise mechanisms, \citet{awan2019benefits} extended the exponential mechanism \citep{mcsherry2007mechanism} to arbitrary Hilbert spaces and showed its application to functional principal component analysis. From the robust noise injection perspective, a heterogeneous noise injection scheme \citep{phan2019heterogeneous} was proposed by assigning different weighted privacy budgets to each coordinate to further improve the robustness of private summaries.

\subsection{Notations and Organization}
The following notations are used throughout the rest of this work and follow standard conventions. For asymptotic notations: $f(n) = O(g(n))$ or $f(n) \lesssim O(g(n))$ means for all $c$ there exists $k>0$ such that $f(n)\leq c g(n)$ for all $n\geq k$; $f(n)\asymp g(n)$ means $f(n) = O(g(n))$ and $g(n) = O(f(n))$; $f(n) = o(g(n))$ means for any $c>0$ there exists $k>0$ such that $f(n)\leq c g(n)$ for all $n\geq k$. 

The rest of the paper is organized as follows. In Section~\ref{sec: preliminary section}, we provide preliminaries on functional summaries and spaces, differential privacy, and a generalization of finite-dimensional embedding mechanisms. In Section~\ref{sec: feasibility section}, we formally propose the ICLP mechanism and establish its feasibility in separable Hilbert spaces (Theorem~\ref{thm: the ICLP mechanism theorem}) and in the space of continuous functions (Theorem~\ref{thm: continuity ICLP theorem}). In Section~\ref{sec: methodology section}, we propose approaches for constructing qualified summaries for the ICLP mechanism and apply them to various statistical applications with utility analysis. Implementation of the mechanism is provided in Section~\ref{sec: algo and implement}. We evaluate the performance of the proposed mechanism on both synthetic datasets and real-world applications in Sections~\ref{sec: simulation section} and \ref{sec: real data section}. Concluding remarks are given in Section~\ref{sec: conclusion section}. Many technical results, including proofs, lemmas, and propositions, are deferred to the Appendix

\section{Preliminaries}\label{sec: preliminary section}

\subsection{Functional Summaries and Spaces}
First, we define the functional summary. Let $\mbH$ be an infinite-dimensional real separable Hilbert space with inner product $\langle \cdot, \cdot \rangle_\mbH$. For a set $\mcX$, we define $\mathcal{D} = \mcX^{n}$ as the collection of all possible $n$-unit datasets, and let $D$ be an element of $\mathcal{D}$. This paper considers the functional summary statistic of interest as an element of $\mbH$, i.e., $f:\mcD \rightarrow \mbH$. 

Next, we briefly introduce the background of random elements in $\mbH$ and some spaces we will work on in this paper. We refer readers to \citet{hsing2015theoretical} for a detailed introduction. A random element $X \in \mbH$ is said to have mean $\mu \in \mbH$ and (linear) covariance operator $C:\mbH \to \mbH$ if 
\begin{equation*}
    \E[\langle X, h \rangle_\mbH] = \langle \mu, h \rangle_\mbH
    \qquad \text{and}
    \qquad
    \Cov(\langle X, h_1 \rangle_\mbH,
    \langle X, h_2 \rangle_\mbH)
    = \langle C h_1, h_2 \rangle_\mbH, 
\end{equation*}
for any $h,h_1,h_2$ in $\mbH$. When $\E \|X - \mu\|_{\mbH}^2 < \infty$, the covariance operator $C$ exists, is self-adjoint, positive semidefinite, and trace class (thus Hilbert-Schmidt and compact) \citep{bosq2000linear,hsing2015theoretical}. According to the spectral theorem for self-adjoint compact operators, $C$ admits the eigen decomposition as
\begin{equation*}
    C(h) = \sum_{j\geq1} \lambda_{j} \langle h, \phi_{j} \rangle_{\mbH} \phi_{j}, \quad \forall h\in \mbH,
\end{equation*}
where $\{ \lambda_j\}_{j\geq1}$ and $\{ \phi_j \}_{j \geq 1}$ are the eigenvalues and eigenfunctions of $C$, respectively.

We define two norms associated with $C$ as
\begin{equation*}
    \left\|h\right\|_{C} = \sqrt{ \sum_{j=1}^{\infty} \frac{\langle h, \phi_j \rangle_{\mbH}^2}{\lambda_j}}
    \quad \textit{and} \quad
    \left\|h\right\|_{1,C} =  \sum_{j=1}^{\infty} \frac{|\langle h, \phi_j \rangle_{\mbH}|}{\sqrt{\lambda_j}}.
\end{equation*}
We denote the two subspaces of $\mbH$ induced by $\|\cdot\|_{C}$ and $\|\cdot\|_{1,C}$ as $\mcH_C = \{h \in \mbH: \|h\|_C < \infty \}$ and $\mcH_{1,C} =  \{h \in \mbH: \|h\|_{1,C} < \infty \}$, respectively.
Note $\left\| \cdot \right\|_{C}$ is the classic Cameron-Martin norm induced by $C$ \citep{bogachev1998gaussian} and $\mcH_C$ is called the Cameron-Martin space of $C$. The $\left\|\cdot\right\|_{1,C}$-norm is analogous to a weighted $\ell^1$-norm. The space $\mcH_{1,C}$ is included in $\mcH_{C}$, i.e., $h\in \mcH_{1,C}$ leads to $h\in \mcH_{C}$.

We also define the power operator of $C$. For any $s\geq 0$, the power operator $C^{s}$ is defined as 
\begin{equation*}
    C^{s}(h) = \sum_{j\geq1} \lambda_{j}^{s} \langle h, \phi_{j} \rangle_{\mbH} \phi_{j}, \quad \forall h\in \mbH,
\end{equation*}
meaning $C^{s}$ shares the same eigenfunctions as $C$ while the eigenvalues are raised to the power of $s$. We define its corresponding power space as \begin{equation*}
    \mcH_{C^{s}} = \left\{ h\in \mbH: \left\|h\right\|_{C^{s}} := \sqrt{ \sum_{j=1}^{\infty} \frac{\langle h, \phi_j \rangle_{\mbH}^2}{\lambda_{j}^{s}}} < \infty \right\}.
\end{equation*}
The space $\mcH_{1,C^{s}}$ and its associated norm $\|\cdot\|_{1,C^{s}}$ follow a similar definition.

\subsection{Differential Privacy}
For a given non-private functional summary statistic $f_{D}$, we denote its private version as $\tilde{f}_D$, which is a random element of $\mbH$ indexed by $D$. We state the definition of differential privacy in terms of conditional distributions \citep{wasserman2010statistical}.
\begin{definition}\label{def: definition of DP}
Let $\tilde{f}_{D}$ be the privatized functional summary of $f_D$. Assume $\{P_{D} : D\in \mathcal{D}\}$ is the family of probability measures over $\Omega$ induced by $\{\tilde{f}_D : D \in \mathcal{D}\}$. We say $\tilde{f}_{D}$ achieves $(\epsilon,\delta)$-DP if for any two adjacent datasets (different in only one record) $D$ and $D'$, and any measurable set $A\in \mathcal{F}$, one has 
\begin{equation}\label{eqn: DP definition}
    P_{D}(A) \leq e^{\epsilon} P_{D'}(A) + \delta.
\end{equation}
In particular, if $\delta = 0$, we say $\tilde{f}_{D}$ achieves $\epsilon$-DP.
\end{definition}
The definition implies that the summaries of two adjacent datasets should have almost the same probability distribution. The privacy budget $\epsilon$ controls how much privacy will be lost while releasing the result, and a small $\epsilon$ implies a higher similarity between $P_{D}$ and $P_{D'}$, and thus increased privacy. Before introducing different DP mechanisms, we first define the global sensitivity of a summary statistic, a central concept of DP \citep{dwork2006calibrating}.
For a functional summary $f: \mcD \rightarrow \mbH$ and a norm $\|\cdot\|$ in the Hilbert space $\mbH$, the global sensitivity of the summary $f_{D}$ with respect to the norm $\|\cdot\|$ is given by 
\begin{equation*}
    \Delta = \sup_{D \sim D'} \| f_{D} - f_{D'}  \|,
\end{equation*}
where $D \sim D'$ means $D$ and $D'$ are adjacent datasets.
Here, the norm $\|\cdot\|$ is typically tailored based on the employed DP mechanism, e.g., $K$-norm for the K-norm mechanism \citep{hardt2010simple,awan2021structure}, Cameron-Martin space norm for the Gaussian mechanism in separable Banach space \citep{mirshani2019formal}. Given that global sensitivity quantifies how much a summary can change with the modification of a single record in the dataset, the additive noise must be calibrated proportionally to the global sensitivity.

\subsection{Finite-Dimensional Representation-wise Laplace}
We introduce the \textit{\underline{F}inite-Dimensional \underline{R}epresentation-wise \underline{L}aplace} (FRL) mechanism, which generalizes almost all current additive noise mechanisms for $\epsilon$-DP that rely on finite-dimensional representations, such as \citet{wang2013efficient,chandrasekaran2014faster,alda2017bernstein,zhang2012functional}, to name a few. Let $\{e_j\}_{j\geq1}$ be an orthonormal basis in $\mbH$. Then, one can approximate the summary using $M$ basis functions, i.e.,
\begin{equation*}
\hat{f}_{D} = \sum_{j= 1}^{M} f_{D j} e_j \quad \textit{with} \quad f_{D j} = \langle f_D, e_j \rangle_{\mbH}.
\end{equation*}
Expanding the functional summaries via a finite basis facilitates dimension reduction so that the classic multivariate i.i.d. Laplace mechanism \citep{dwork2014algorithmic} can be implemented to privatize the coefficient vector $( f_{D 1},\cdots, f_{D M})$. Specifically, the privatized summary can be expressed as
\begin{equation*}
    \tilde{f}_{D} = \sum_{j= 1}^{M} (f_{D j} + Z_{j}) e_j, \quad \text{with} \quad Z_{j}\stackrel{i.i.d.}{\sim} Lap(0, \Delta / \epsilon)
\end{equation*}
where $\Delta$ is the global sensitivity in $\ell^{1}$-distance. The FRL mechanism privatizes the functional summary $\hat{f}_{D}$ without regard to the varying importance of different components. Although some components are more crucial for the estimation, the FRL mechanism treats all components equally during the privatization process, thereby reducing the utility of the privatized summary. Additionally, the truncation level $M$ controls the trade-off among variance, bias, and privacy. This mechanism forces one to either introduce more noise or accept higher bias when more components are required to deal with complex functional summaries.

The $m$-dimensional coefficient vector can also be privatized using the $K$-norm mechanism \citep{hardt2010geometry,awan2021structure}, by perturbing the coefficients through a multivariate random variable in $\mbR^{M}$, whose density is proportional to $\exp\{-\epsilon\|\cdot\|_{K}\}$ for a given $K$-norm in $\mbR^{M}$. Utilizing the $\ell^{1}$-norm results in the aforementioned multivariate i.i.d. Laplace mechanism (the FRL mechanism). However, as noted in \citet{awan2021structure}, when $M$ is large, determining the optimal $K$-norm is often nontrivial, and sampling these multivariate random variables can be challenging. Due to these challenges and the prevalence of all existing $\epsilon$-DP mechanisms employing finite-dimensional representation with i.i.d. Laplace noise for functional summaries, our discussion primarily focuses on the FRL mechanism. We leave the exploration of using other $K$-norms with finite-dimensional representation for privatizing functional summaries for future studies.

\section{The Independent Component Laplace Process Mechanism}\label{sec: feasibility section}
In this section, we first formally define the Independent Component Laplace Process and then propose an additive noise mechanism called \textit{the ICLP mechanism}. Specifically, to achieve $\epsilon$-DP, we will release the privatized summary that takes the form of $\tilde{f}_{D} = f_{D} + \sigma Z$, where $\sigma$ is a positive scalar and $Z$ is an ICLP noise. Initially, we assume $f_D$ lies in a real separable Hilbert space $\mbH$ and establish the $\epsilon$-DP guarantee. We then show that privacy protection can also hold for the space of continuous functions (which is not a Hilbert space) under certain assumptions on the covariance operator. The proofs of all the theorems can be found in Appendix \ref{apd: proofs for ICLP}.

\subsection{Independent Component Laplace Process}
The proposed random element is motivated by \citet{mirshani2019formal}, who achieved $(\epsilon,\delta)$-DP for functional summaries in Banach spaces. Formally, their additive noise mechanism can be expressed as $\tilde{f}_D = f_D + \sigma Z$, where $Z \sim GP(0, C)$ is a centered Gaussian process with covariance operator $C$. There is a dual perspective of this mechanism. By applying the Karhunen-Lo\'eve Theorem \citep{kosambi2016statistics}, the mechanism is equivalent to 
\begin{equation}\label{eqn: KL expansion form}
    \tilde{f}_{D} = f_D + \sigma Z = \sum_{j=1}^{\infty} \left( \left\langle f_D, \phi_j \right\rangle + \sigma \left\langle Z, \phi_j \right\rangle \right) \phi_j,
\end{equation}
where $\{\lambda_j\}_{j\geq 1}$ and $\{\phi_j\}_{j\geq 1}$ are the eigenvalues and eigenfunctions of $C$ and $\{\langle Z, \phi_j \rangle\}_{j\geq 1 }$  are independent Gaussian random variables with zero mean and variance $\lambda_{j}$. This decomposition indicates that the mechanism perturbs each coefficient with independent Gaussian random variables. Unfortunately, the existing Laplace process cannot play a role analogous to the Gaussian process under such a decomposition, as it is an elliptical distribution. It has been proven that no elliptical distribution can achieve $\epsilon$-DP in infinite-dimensional spaces. Specifically, in an infinite-dimensional space, adding any elliptical distribution is equivalent to adding noise from a randomly scaled Gaussian process, which satisfies only the weaker notion of $(\epsilon,\delta)$-DP. We refer readers to Theorem 4 of \citet{reimherr2019elliptical} for more details. Motivated by this dual perspective via the Karhunen-Lo\'eve expansion and the fact that the most widely used additive noise mechanism for $\epsilon$-DP in the univariate case is the Laplace mechanism, we consider using independent Laplace random variables with heterogeneous variances in the decomposition (\ref{eqn: KL expansion form}). This is equivalent to perturbing the functional summary with a particular random element defined as follows.
\begin{definition}\label{def: definition of ICLP}
Let $X$ be a random element in $\mbH$ with $\E \|X\|_{\mbH}^2 < \infty$ and $C$ be its covariance operator. Denote the eigenvalues and eigenfunctions of $C$ as $\{\lambda_{j}\}_{j\geq 1}$ and $\{\phi_{j}\}_{j\geq 1}$. We say $X$ is an Independent Component Laplace Process (ICLP) with mean $\mu$ if it admits the following decomposition 
\begin{equation}\label{eqn: ICLP KL-decomposition}
     X = \mu + \sum_{j=1}^\infty \sqrt{\lambda_j} Z_{j} \phi_{j},
\end{equation}
where $Z_j$ are i.i.d. Laplace random variables with zero mean and variance $1$.
\end{definition}
\begin{remark}
If the objective is to achieve $\epsilon$-DP, alternative i.i.d. sequences of random variables $\{Z_{j}\}_{j\geq 1}$ with zero mean and unit variance, characterized by a well-constructed density, are technically viable. However, this requires that the density tail of these random variables be calibrated ``just right'' to comply with $\epsilon$-DP. Typically, their density tails should be as heavy, or closely similar, to the Laplace distribution. Otherwise, using light-tailed random variables, like Gaussian, will cause the probability inequality (\ref{eqn: DP definition}) of $\epsilon$-DP to fail for sets in the tails.
\end{remark}
The collection of square-integrable random elements of $\mbH$ is itself a Hilbert space with inner product $ \E \langle X, Y \rangle_\mbH$. The following theorem states that if a random element $X$ is defined via the infinite sum decomposition in Definition~\ref{def: definition of ICLP}, it is still well-defined in $\mbH$.
\begin{theorem}\label{thm: existence of iclp}
For a given non-negative decreasing real sequence $\{\lambda_j\}_{j\geq 1}$ that is summable, and an orthonormal basis $\{\phi_j\}_{j\geq 1}$ for $\mbH$, the random element $X$ defined via (\ref{eqn: ICLP KL-decomposition}) is well-defined within $\mbH$.
\end{theorem}

\subsection{Feasibility in Separable Hilbert Spaces}\label{hilbertspace subsection}

When the summary $f_{D}$ of interest is infinite-dimensional, it turns out that the summary $f_{D}$ depends heavily on the structure of the random element $Z$ in the additive noise mechanism. Otherwise, it is possible to make the global sensitivity infinite, and thus, no finite amount of noise would be able to achieve DP. For example, in \citet{mirshani2019formal}, the privatized summary is $\tilde{f}_D = f_D + \sigma Z$, where $Z$ is a zero-mean Gaussian process with covariance operator $C$. It has been proved that the summary $f_{D}$ must be ``compatible'' with the Gaussian process $Z$, i.e., $f_{D}-f_{D'}$ lies in the Cameron-Martin space of $Z$ for any adjacent datasets $D, D'$, to achieve $(\epsilon,\delta)$-DP, or no finite $\sigma$ will make the mechanism satisfy $(\epsilon,\delta)$-DP. Given that our problem setting is also infinite-dimensional, a similar analysis is needed for the ICLP mechanism. In this section, we will investigate the feasibility of the ICLP mechanism, i.e., identifying the specific conditions under which there always exists a finite $\sigma$ such that the privatized summary $f_D + \sigma Z$ via the ICLP mechanism achieves $\epsilon$-DP.

To investigate the feasibility of a randomized mechanism for $\epsilon$-DP, one can start with the equivalence or orthogonality of probability measures. As discussed in \citet{awan2019benefits} and \citet{reimherr2019elliptical}, the probability measures induced by an $\epsilon$-DP mechanism are necessarily equivalent (though this is not sufficient for DP) in a probabilistic sense; otherwise, it is impossible to achieve DP if the measures are orthogonal. More specifically, if the mechanism produces a private summary $\tilde{f}_{D}$ that is probabilistically orthogonal to $\tilde{f}_{D'}$, i.e., there exists a $A\in \mathcal{F}$ such that $P_{D}(A) = 0$ and $P_{D'}(A) = 1$, then the mechanism cannot be DP since $f_{D}$ and $f_{D'}$ can be distinguished with probability one on $A$. In the following, we use this perspective to develop the feasibility of the ICLP mechanism. Denote the probability measure family induced by the ICLP mechanism as $\{P_{D}:D\in \mathcal{D}\}$. In the following theorem, we provide necessary and sufficient conditions for pairwise equivalence in $\{P_{D}:D\in \mathcal{D}\}$.
\begin{theorem}[Equivalence of ICLP probability measures]\label{Thm: Equivalence of ICLP probability measure}
Let $D, D' \in \mathcal{D}$ be two adjacent datasets, $\tilde{f}_{D}, \tilde{f}_{D'}$ be the privatized summaries based on the ICLP mechanism. Denote the corresponding probability measures over $\mathbb{H}$ as $P_{D}$ and $P_{D'}$, and the covariance operator of ICLP as $C$. Then $P_{D}$ and $P_{D'}$ are equivalent if and only if 
\begin{equation}\label{formula of H_K}
    f_{D} - f_{D'} \in \mcH_{C} = \left\{  h\in \mathbb{H} : \|h\|_{C} < \infty \right\}.
\end{equation}
\end{theorem}
Theorem~\ref{Thm: Equivalence of ICLP probability measure} shows that if the difference between $f_{D}$ and $f_{D'}$ resides in the Cameron-Martin space of $C$, then the probability family will be pairwise equivalent. An analogous result for the equivalence of elliptical distributions appears in Theorem 2 of \citet{reimherr2019elliptical}, even though the ICLP is not an elliptical distribution. However, it turns out that, unlike elliptical distributions, pairwise equivalence is not enough for the ICLP mechanism to achieve $\epsilon$-DP. To see the reason behind this, one must consider the density of $P_D$ in $\mbH$. Since there is no common base measure in $\mathbb{H}$ that plays the same role as the Lebesgue measure in $\mbR^{d}$, it is more complicated to consider the density in $\mathbb{H}$.
Fortunately, we are adding the same type of noise to the functional summaries. Therefore, we only need the density as the Radon-Nikodym derivative of $P_{D}$ with respect to $P_{0}$, where $P_{0}$ is the probability measure induced by $\sigma Z$. 
\begin{theorem}[Density of ICLP]\label{thm: density}
Let $P_h$ and $P_{0}$ be the probability measures induced by $\{ h + \sigma Z \}$ and $\sigma Z$ respectively. Suppose $h \in \mcH_{1,C}$, then the Radon–Nikodym derivative of $P_{D}$ with respect to $P_{0}$ is given by
\begin{equation}\label{eqn: formula of density}
    \frac{d P_h }{d P_{0} }(z) = \exp\left\{ -\frac{1}{\sigma} \left( \left\|z - h\right\|_{1,C} - \left\| z \right\|_{1,C} \right) \right\},
\end{equation}
$P_0$ almost everywhere and is unique. 
\end{theorem}
Now, we are ready to show why the condition (\ref{formula of H_K}) is insufficient for $\epsilon$-DP. Indeed, even though $f_{D} - f_{D'} \in \mcH_{C}$ guarantees the pairwise equivalence between $P_{D}$ and $P_{D'}$, it does not guarantee the density in Equation (\ref{eqn: formula of density}) is well-defined and thus cannot be upper bounded, which is a requirement for $\epsilon$-DP however.
Meanwhile, $\mcH_C$ is enough for $(\epsilon,\delta)$-DP with Gaussian process with covariance $C$ \citep{mirshani2019formal} since it allows densities to be unbounded up to a set with $P_0$ measure less than $\delta$. In the following theorem, we will show the appropriate space in which $f_{D} - f_{D'}$ should reside is the subspace of $\mcH_{1, C}$.

\begin{theorem}[Impossibility of The ICLP Mechanism]\label{thm: gap theorem}
Under the same conditions of Theorem~\ref{Thm: Equivalence of ICLP probability measure}, let $\mcH_{1,C} = \left\{  f\in \mathbb{H} : \|f\|_{1,C} < \infty \right\}$ be a subspace of $\mcH_{C}$ and if 
\begin{equation*}
    f_{D_1} - f_{D_2}  \in \mcH_{C} \setminus \mcH_{1,C},
\end{equation*}
then there is no $\sigma \in \mathbb{R}^{+}$ such that the ICLP mechanism, $\tilde{f}_{D} = f_{D} + \sigma Z$, satisfies $\epsilon$-DP.
\end{theorem}
Indeed, if $f_{D}$ resides in the gap between $\mcH_{C}$ and $\mcH_{1, C}$, the sensitivity of $f_{D}$ will be infinite, and there is no possibility to calibrate the ICLP noise with any finite $\sigma$ to achieve $\epsilon$-DP. Now, with the proper space in Theorem \ref{thm: gap theorem} and the feasible density in Theorem \ref{thm: density}, we can establish the ICLP mechanism formally.
\begin{theorem}[The ICLP Mechanism]\label{thm: the ICLP mechanism theorem}
Let $f_{D}$ be the functional summary and $Z$ be an ICLP with covariance operator $C$. Define the global sensitivity of the ICLP mechanism as
\begin{equation}\label{formula of GS}
    \Delta = \sup_{D \sim D'} \| f_{D} - f_{D'} \|_{1,C} \quad \textit{and} \quad \sigma = \Delta / \epsilon.
\end{equation}
Then the privatized version of $f_{D}$, $\tilde{f}_{D} = f_{D} + \sigma Z$, achieves $\epsilon$-DP.
\end{theorem}


\subsection{Example of $\mcH_{C}$ and $\mcH_{1,C}$ for Smooth Functions}
In this section, we provide a concrete example of the spaces $\mcH_{C}$ and $\mcH_{1,C}$, in which the smoothness of the elements in these spaces is of primary interest. Consider the Hilbert space $\mbH$ as $L^{2}(\mcX)$ where $\mcX \subseteq \mbR^{d}$. For the ICLP covariance function $C$ defined over $\mcX\times \mcX$, the eigenfunctions $\{\phi_{j}\}_{j\geq 1}$ are an orthonormal basis of $L^{2}(\mcX)$. Suppose the eigenvalues of $C$ decay polynomially with order $\beta$, i.e., $\lambda_{j}\asymp j^{-\beta}$. Then the space $\mcH_{C}$ and $\mcH_{1,C}$ thus can be expressed as 
\begin{equation*}
\begin{aligned}
    & \mcH_{C} = \left\{ f=\sum_{j=1}^{\infty} f_{j} \phi_{j} \in L^{2}(\mcX): \sum_{j=1} j^{2\beta} f_{j}^2 < \infty \right\}, \\
    & \mcH_{1,C} = \left\{ f=\sum_{j=1}^{\infty} f_{j} \phi_{j} \in L^{2}(\mcX): \sum_{j=1} j^{\beta} |f_{j}| < \infty \right\}.
\end{aligned}
\end{equation*}
If $\{\phi_{j}\}_{j\geq 1}$ are the Fourier basis, $\mcH_{C}$ corresponds to a Sobolev space of order $\beta$, containing functions with weak derivatives up to order $\beta$ that are $L^{2}$-integrable. $\mcH_{1,C}$ corresponds to a H\"{o}lder space of order $\beta$, consisting of those functions with continuous derivatives up to order $\lfloor \beta\rfloor$, and the $\lfloor \beta\rfloor$-th derivative is $\beta - \lfloor \beta\rfloor$ H\"{o}lder continuous. We refer readers to Section 2.3 of \citet{yang2017frequentist} for further details.

\subsection{Extensions To the Space of Continuous Functions}\label{extension subsection}

Theorem \ref{thm: the ICLP mechanism theorem} implies that the ICLP mechanism provides privacy protection for a wide range of infinite-dimensional functional objects in separable Hilbert spaces. The DP \textit{post-processing inequality} \citep{dwork2014algorithmic} is a fundamental property for functional summaries since one may only be practically interested in a few scalar summaries. However, the post-processing inequality only applies to measurable mappings. If $\mbH = L^{2}([0,1])$, then this eliminates the possibility of releasing point-wise evaluations of the functional summary since such mappings are not measurable operators in $L^{2}([0,1])$. Therefore, in this section, we extend the ICLP mechanism to the space of continuous functions, i.e., $\mcC(T)$ with $T$ a compact set over $\mathbb{R}^{d}$, where such operators are measurable (and thus protected). 
We show that the ICLP, under mild conditions, is also in $\mcC(T)$.

\begin{theorem}[Feasibility in the Space of Continuous Functions]\label{thm: continuity ICLP theorem}
Let $C: T \times T \to \mbR$ be a symmetric, positive definite, bivariate function over compact domain $T$. If $C$ is $\alpha$-H\"{o}lder continuous in each coordinate, i.e., there exists a positive constant $M_C$, and $\alpha \in (0,1]$ such that $\left| C\left(t_1,s\right) - C\left(t_2,s\right) \right| \leq M_C \left| t_1 - t_2 \right|^{\alpha}$,
then there exists an ICLP, $Z$, with covariance function $C$ and a modification $\tilde{Z} : T \times \Omega \rightarrow \mbR$ of $Z$ that is a continuous process, such that
\begin{enumerate}
    \item  $\tilde{Z}$ is sample continuous, i.e., $\forall \omega \in \Omega$, $\tilde{Z}_{\omega}(t)$ is continuous with respect to $t\in T$;
    
    \item For any $t\in T$, $P( \tilde{Z}(t) = Z(t)) = 1$.
\end{enumerate}
Meaning that there exists a stochastic process in $\mcC(T)$ equally distributed as the ICLP except on a zero-measure set. 
\end{theorem}
All the ICLP mechanism results for $\mbH$ in Section \ref{hilbertspace subsection} are now applicable to functional summaries in $\mcC(T)$. Furthermore, the point-wise evaluation is now a measurable operation and thus is protected.  We also note that the proof of Theorem \ref{thm: continuity ICLP theorem} is not just a standard result from stochastic processes but relies heavily on the structure of the ICLP.

\section{Qualified Summary Obtainment and Privacy-Preserving Tasks}\label{sec: methodology section}
In this section, we present different approaches to constructing non-private summaries for the ICLP mechanism. Specifically, these constructions must ensure that the difference of functional summaries, $f_{D} - f_{D'}$, lies in $\mcH_{1,C}$ for any adjacent datasets $D$ and $D'$, thus qualifying the summaries for the ICLP mechanism. After introducing these approaches, we also apply the ICLP mechanism to achieve privacy protection in different statistical estimation problems with corresponding statistical analysis.

\subsection{Generalized Obtainment of Qualified Summaries} \label{ERM approach section}
Based on Theorems \ref{thm: gap theorem} and \ref{thm: the ICLP mechanism theorem}, to achieve $\epsilon$-DP via the ICLP mechanism, the difference of non-private summaries calculated from any adjacent datasets $D$ and $D'$ should reside in $\mcH_{1,C}$. We call a summary $f$ that satisfies such conditions a qualified summary. However, constructing such $f$ directly is a challenging task. It is easier to address the problem by restricting the individual functional summary $f_{D}$ residing in $\mcH_{1,C}$ for any $D\in \mcD$, which automatically leads to $f_{D} - f_{D'} \in \mcH_{1,C}$. Therefore, we leverage regularized ERM to obtain qualified summaries $f_{D}$ such that $f_{D}\in \mcH_{1,C}$. 

Formally, let $L(f,D): \mbH \times \mathcal{D} \to \mathbb{R}$ be a loss function. The \textit{ICLP with Absolute Regularization} (ICLP-AR) estimator is defined as follows,
\begin{equation}\label{Lasso-based-Strategy problem}
    (\text{ICLP-AR}): \quad \hat{f}_{D} = \underset{f\in \mathbb{H}}{\operatorname{argmin}} \left\{ L(f,D) + \psi \left\| f \right\|_{1,C^{\eta}} \right\} \quad \textit{for} \ \eta \geq 1,
\end{equation}
where $C^{\eta}$ is the power kernel of $C$ that shares the same eigenfunctions as $C$ while the eigenvalues are raised to $\lambda_{j}^{\eta}$ and $\psi$ is the regularization parameter.

The benefits of using a power kernel $C^{\eta}$ are twofold. First, the space corresponding to $\|\cdot\|_{1,C^{\eta}}$ is a subspace of $\mcH_{1,C}$, guaranteeing that $\hat{f}_{D} \in \mcH_{1,C}$. Second, it allows more flexibility to control the regularity (usually smoothness) of the constructed functional summaries. Later on, we will see that even though $\eta=1$ is a natural setting, setting $\eta>1$, i.e., constructing a slightly over-smoothing summary, can be helpful for utility and even make privacy error negligible compared to estimation error. However, as we will see in Section \ref{sec: mean function protection}, there are some serious drawbacks to using the $\|\cdot\|_{1,C}$-norm regularization.

Therefore, we consider restricting the functional summary in the power space of $\mcH_{C}$, i.e., $\mcH_{C^{\eta}}$ and using $\|\cdot\|_{C^{\eta}}$-norm regularization in regularized ERM as our final strategy, which turns out to work quite well theoretically and practically. Formally, for a given $\eta > 1$, by the Cauchy-Schwarz inequality we have
\begin{equation*}
    \|h\|_{1,C} = \sum_{j=1}^{\infty} \frac{|h_j|}{\sqrt{\lambda_j}} = \sum_{j=1}^{\infty} \frac{|h_j|}{\lambda_j^{\frac{\eta}{2}}}\lambda_j^{\frac{\eta-1}{2}} \leq   \left\| h \right\|_{C^{\eta}} \sqrt{\operatorname{trace}(C^{\eta-1})}.
\end{equation*}
Therefore, by taking $\eta>1$ such that $C^{\eta-1}$ is a trace-class operator, we obtain the functional summary via the following \textit{ICLP with Quadratic Regularization} (ICLP-QR) strategy,
\begin{equation}\label{ICLP-QR problem}
    \text{(ICLP-QR)}: \quad \hat{f}_{D} = \underset{f\in\mbH}{\operatorname{argmin}} \left\{ L(f,\mathcal{D}) + \psi \left\| f\right\|_{C^{\eta}}^2 \right\} \quad \textit{for}  \ \eta > 1.
\end{equation}
Since the power space $\mcH_{C^{\eta}}$ is an RKHS and $\|\cdot\|_{C^{\eta}}$ takes quadratic form, we name this approach ICLP-QR.
Here, $\eta$ is strictly greater than $1$ leads to $\mcH_{C^{\eta}}\subseteq \mcH_{1,C} \subseteq \mcH_{C}$, which ensures the feasibility of the ICLP mechanism. Similar to ICLP-AR, the $\eta$ in ICLP-QR also plays a role in balancing the utility and the privacy of the functional summary.

\subsection{Protection for Mean}\label{sec: mean function protection}
We consider the problem of privatizing the mean summary. Assume $X_{1},\cdots,X_{n}$ are i.i.d. random elements drawn from an arbitrary real separable Hilbert space $\mbH$ with mean element $\E X_{i} = \mu_{0} \in \mbH$. Our goal is to release a private estimator for the true mean $\mu_{0}$ that satisfies $\epsilon$-DP. When using the FRL mechanism, one can start with the sample mean $\hat{\mu}_{D} = \frac{1}{n}\sum_{i=1}^{n}X_{i}$, which is an unbiased estimator of $\mu_{0}$. For the ICLP mechanism, we use regularized ERM with the square loss to obtain qualified non-private summaries, i.e.,
\begin{equation}\label{eqn:empirical minimium}
    \hat{\mu}_{D} = \underset{\theta\in \mbH }{\operatorname{argmin}} \left\{ \frac{1}{n}\sum_{i=1}^{n}\left\| X_i - \theta \right\|_{\mbH}^{2} + \psi P(\theta) \right\},
\end{equation}
with $P(\theta) = \left\| \theta \right\|_{1,C^{\eta}}$ or $\left\| \theta \right\|_{C^{\eta}}^2$ corresponding to ICLP-AR and ICLP-QR respectively.

To obtain the global sensitivity and utility analysis for the proposed strategies, we first state some standard assumptions in the DP literature regarding the norm of the observed data and the eigenvalue decay rate of $C$.
\begin{assumption}[Boundedness]\label{assmption: bounding norm}
Assume for any sample path $X$, its $\mbH$-norm is bounded by $\tau$, i.e., $\|X\|_{\mbH}\leq \tau$. 
\end{assumption}
The bounded norm assumption is commonly used in the DP literature, primarily to ensure finite global sensitivity. This assumption is often adapted to align with specific DP paradigms and mechanisms. For instance, in employing the Gaussian mechanism to attain $(\epsilon,\delta)$-DP, it is customary to assume that the $\ell^{2}$-norm of the data is bounded, in line with the global sensitivity being assessed via $\|\cdot\|_{\ell^{2}}$. Conversely, for the Laplace mechanism, which aims for $\epsilon$-DP, the focus shifts to the $\ell^{1}$-norm, with global sensitivity gauged through $\|\cdot\|_{\ell^{1}}$. In cases where a more general norm in $\mbR^{d}$ is used, such as in the exponential mechanism, the data is presumed to have such a finite general norm. See the mean estimation example in \citet{reimherr2019kng}. 

In the context of the ICLP mechanism, where global sensitivity is evaluated under $\|\cdot\|_{1,C}$, it seems logical to posit that $\|X\|_{1,C}\leq \tau$ for some finite $\tau$. However, the $\|\cdot\|_{1,C}$-norm is intrinsically linked to $C$, dependent on the eigenvalues and eigenfunctions of the covariance used in ICLP. This reliance implies that assuming $\|X\|_{1,C}\leq \tau$ could narrow the ICLP mechanism's applicability. Practically, it would mean that feasible $X_{i}$ should be drawn from $X$ whose covariance eigenfunctions align with those of ICLP's covariance $C$, a condition often unverifiable in practice. Therefore, we consider a more general and relaxed boundedness assumption by assuming $\|X\|_{\mbH}\leq \tau$, which is more likely to be met in most practical applications.

\begin{assumption}[Eigenvalue Decay Rate (EDR)]\label{assump: EDR}
    Suppose the eigenvalue decay rate of $C$ is $\beta > 1$, i.e., there exist constants $c_{1}$ and $c_{2}$ such that 
    \begin{equation*}
        c_{1} j^{-\beta} \leq \lambda_{j} \leq c_{2} j^{-\beta},\quad \forall i = 1,2,\cdots.
    \end{equation*}
\end{assumption}
Note that the eigenvalues $\lambda_{j}$ and EDR are only determined by the ICLP covariance $C$. The polynomial eigenvalue decay rate assumption is standard in the non-parametric literature. For example, if $C$ satisfies the Sacks–Ylvisaker conditions \citep{sacks1966designs,sacks1968designs,sacks1970designs} of order $s$, then $\lambda_{j}\asymp j^{-2(s+1)}$. If setting $C$ equal to the reproducing kernel of the univariate Sobolev space $\mathcal{W}_{2}^{m}([0,1])$ results in $\lambda_{j}\asymp j^{-2m}$, see \citet{micchelli1979design} for more instances. 

In the following theorems, we derive the closed form of the estimators and provide their global sensitivity analysis for the FRL, ICLP-AR, and ICLP-QR mechanisms.
\begin{theorem}[Global Sensitivity Analysis]\label{thm: GS analysis for mean protection}
Suppose Assumption \ref{assmption: bounding norm} holds, then
\begin{enumerate}
    \item (FRL) Suppose the functional summary used in FRL is a truncated sample mean function, i.e., $\hat{\mu}_{D} = \sum_{j=1}^{M}\langle\bar{X},\phi_{j} \rangle_{\mbH}\phi_{j}$, then
    \begin{equation*}
        \Delta = \underset{D,D'}{\operatorname{max}} \| \hat{\mu}_{D} - \hat{\mu}_{D'} \|_{\ell^{1}} \leq \frac{2M\tau}{n}.
    \end{equation*}

    \item (ICLP-AR) The solution of ICLP-AR in (\ref{eqn:empirical minimium}) is 
    \begin{equation}\label{eqn: ICLP-AR closed form}
        \hat{\mu}_{D} = \sum_{j=1}^{\infty}s_{\psi, 2\lambda_j^{\eta/2}}\left( \left\langle \bar{X},\phi_j \right\rangle_{\mbH} \right) \phi_{j},
    \end{equation}
    for all $\eta\geq 1$ and $s_{a,b}(x) = \operatorname{sgn}(x)\left( |x| - a/b\right)^{+}$ is the soft thresholding function with threshold $a/b$. Then, there exists an integer $J^{*}$ such that the global sensitivity of $\hat{\mu}_{D}$ satisfies
    \begin{equation*}
        \sup_{D \sim D'} \| \hat{\mu}_{D} - \hat{\mu}_{D'} \|_{1,C} \leq \frac{2\tau}{n} \sum_{j = 1}^{J^{*}} \lambda_j^{-\frac{1}{2}}.
    \end{equation*}
    
    \item (ICLP-QR) The solution of ICLP-QR in (\ref{eqn:empirical minimium}) is 
    \begin{equation}\label{eqn: ICLP-QR est form}
        \hat{\mu}_{D} = \sum_{j=1}^{\infty} \frac{\lambda_j^{\eta}}{\lambda_j^{\eta} + \psi} \left\langle \bar{X},\phi_j \right\rangle \phi_j,
    \end{equation}
    for all $\eta > 1 + \beta^{-1}$.
    Then, the global sensitivity of $\hat{\mu}_{D}$ satisfies
    \begin{equation*}
        \sup_{D \sim D'} \| \hat{\mu}_{D} - \hat{\mu}_{D'} \|_{1,C} \leq \frac{2\tau}{n} \sum_{j=1}^{\infty}\left( \frac{\lambda_j^{\eta-\frac{1}{2}}}{\lambda_j^{\eta} + \psi} \right).
    \end{equation*}
\end{enumerate}
\end{theorem}
\begin{remark}
The integer $J^{*}: = \operatorname{min} \{ j \geq 1: \tau \leq \psi/2\lambda_j^{\eta/2} \}$ in the ICLP-AR estimator can indeed be viewed as a truncation number as the coefficients after $J^{*}$ will be shrunk to $0$, i.e., the summation in (\ref{eqn: ICLP-AR closed form}) is indeed finite. The upper bound for global sensitivity is based on the fact that, in the worst-case scenario, the coefficients are not shrunk to zero, and thus, the soft thresholding adjustments are canceled out. Therefore, unfortunately, the ICLP-AR estimator does not produce a better sensitivity than the FRL approach, while the soft thresholding introduces extra bias into the summary. On the other hand, the coefficients of the ICLP-QR estimator (\ref{eqn: ICLP-QR est form}) will not be shrunk exactly to zero. Hence, one is able to perturb the functional summary with the truly infinite-dimensional ICLP. 
\end{remark}
We now analyze the error of the non-private summary $\hat{\mu}_{D}$ and the private summary $\tilde{\mu}_{D}$. We call $\E\|\hat{\mu}_{D} - \mu_{0}\|_{\mbH}^2$ the estimation error and the quantity $\E\|\tilde{\mu}_{D} - \hat{\mu}_{D}\|_{\mbH}^2$ the privacy error. We also call the mean square error (MSE) of $\tilde{\mu}_{D}$, $\E\|\tilde{\mu}_{D} - \mu_{0}\|_{\mbH}^2$, the privacy-estimation error, which represents the amount of error in estimating the population mean by the private summary.

\begin{theorem}\label{thm: total error for mean protection}
    Suppose $X_{i}$ are i.i.d. observations drawn from population $X$ with mean function $\mu_{0}$, and Assumptions~\ref{assmption: bounding norm} and \ref{assump: EDR} hold. We also assume $\mu_{0}\in \mcH_{C^{\eta}}$ for some $\eta > 1 + \beta^{-1}$. Let $\hat{\mu}_{D}$ be non-private estimators under different approaches and $\tilde{\mu}_{D}$ as their private version. Then
    \begin{equation*}
        \E \left\| \tilde{\mu}_{D} - \mu_{0} \right\|_{\mbH}^2 =  \left\{ \begin{aligned}
        &O\left(\frac{8M^3\tau^2}{n^2 \epsilon^2} + \frac{1}{n} + M^{-\eta\beta} \right), & \text{(FRL)} \\
        &O\left(\frac{4\tau^2}{n^2 \epsilon^2} (J^{*})^{ \beta + 2} + \frac{1}{n} + \psi \right), & \text{(ICLP-AR)}\\
        &O\left(\frac{8\tau^2}{n^2 \epsilon^2}\text{tr}(C^{\eta - 1}) \psi^{-\frac{2}{\eta}(\frac{\beta+2}{2\beta})} + \frac{1}{n} + \psi \right). & \text{(ICLP-QR)}
    \end{aligned} \right.
    \end{equation*}
\end{theorem}
The first term in the privacy-estimation error is the privacy error, while the last two terms constitute the estimation error. Notably, the optimal estimation error rate for mean function estimation in $\mbH$ is $O(n^{-1})$. By tuning the regularization parameters in different mechanisms, one can ensure that the privacy error either matches or is of a lower order than the estimation error. This tuning allows the private summary to perform as well as the non-private one in terms of error rate. In the following corollary, we provide the optimal order for these regularization parameters such that the privacy error is $O(n^{-1})$ or $o(n^{-1})$.

\begin{corollary}\label{col: bounds for mean protection under l2 norm}
    Under the same conditions as Theorem~\ref{thm: total error for mean protection},
    \begin{enumerate}
        \item (FRL) Setting $n^{\frac{1}{\eta\beta}} \lesssim M \lesssim n^{\frac{1}{3}}$ leads to
        \begin{equation*}
            \E \left\| \tilde{\mu}_{D} - \mu_{0} \right\|_{\mbH}^2 = O(n^{-1}).
        \end{equation*}
        \item (ICLP-AR) Setting $(n \epsilon^2)^{-\frac{\eta \beta}{2(\beta + 2)}}\lesssim \psi \lesssim n^{-1}$ with $\eta \geq 2(1 + 2 \beta^{-1})$ leads to 
        \begin{equation*}
            \E \left\| \tilde{\mu}_{D} - \mu_{0} \right\|_{\mbH}^2 = O(n^{-1}).
        \end{equation*}
        In particular, setting $\psi \asymp n^{-1}$ and $\eta > 2(1 + 2\beta^{-1})$, 
        \begin{equation*}
            \E \left\| \tilde{\mu}_{D} - \hat{\mu}_{D} \right\|_{\mbH}^2 = o(n^{-1}).
        \end{equation*}

        \item (ICLP-QR) Setting $(n \epsilon^2)^{-\frac{\eta \beta}{\beta + 2}}\lesssim \psi \lesssim n^{-1}$ with $\eta \geq 1 + 2 \beta^{-1}$ leads to 
        \begin{equation*}
            \E \left\| \tilde{\mu}_{D} - \mu_{0} \right\|_{\mbH}^2 = O(n^{-1}).
        \end{equation*}
        In particular, setting $\psi \asymp n^{-1}$ and $\eta > 1 + 2\beta^{-1}$, 
        \begin{equation*}
            \E \left\| \tilde{\mu}_{D} - \hat{\mu}_{D} \right\|_{\mbH}^2 = o(n^{-1}).
        \end{equation*}
    \end{enumerate}
\end{corollary}
Corollary \ref{col: bounds for mean protection under l2 norm} suggests that by optimally choosing the finite-dimensional number $M$ in the FRL mechanism or the regularization parameter $\psi$ in ICLP-QR, the privacy error will not dominate the privacy-estimation error, making the privacy-estimation error match the optimal estimation error of $O(n^{-1})$. Additionally, in ICLP-QR, slightly oversmoothing via regularization (choosing $\eta > 1 + 2\beta^{-1}$) results in the privacy error being a lower order of the estimation error, making it asymptotically negligible. This phenomenon is referred to as ``free privacy'' since privatizing the summaries will not affect their statistical performance. Similarly, ICLP-AR can also achieve the optimal error rate $O(n^{-1})$ and even gain ``free privacy'' like ICLP-QR. However, ICLP-AR requires higher oversmoothing to achieve the optimal rate compared to ICLP-QR. This could be attributed to our error analysis using the largest integer $J^{*}$ (beyond which all coefficients are zero), representing the worst-case scenario. In real-world applications, the actual value of $J^{*}$ is usually smaller than what we use in our analysis, leading to a performance that might be slightly worse than that of the FRL mechanism, see Section \ref{sec: simulation section}. However, it's important to note that the true value of $J^{*}$ is not analyzable within the current framework.

This section primarily focuses on the mean estimation and privacy protection problem, a common problem in the functional data analysis where the samples $X_{i}$ are random functions in a function space \citep{rice1991estimating,cai2011optimal}. Although this problem is relatively simple, its approaches and analysis can be methodologically extended to other problems where estimations can be boiled down to estimating the average functional statistic. For example, in functional data analysis, the covariance function estimation is essentially the average of the sample $\{(X_{i}(t) - \bar{X}(t))(X_{i}(s) - \bar{X}(s))\}_{i=1}^{n}$; estimating the coefficient in the function-on-scalar linear regression model is essentially the average of the sample $\{X_{i} Y_{i}(t)\}_{i=1}^{n}$. Given the inherently repetitive nature of these problems, we only provide the analysis of the mean protection and refer readers to Appendix~\ref{apd: extension of mean function} for more detailed discussions on the parallels between these problems.

\subsection{Beyond Mean Protection}
In this part, we demonstrate how the ICLP mechanism is not only feasible for functional summaries but also can be applied to more general learning problems where the summary of interest is a function.

\paragraph{Kernel Density Estimation.}
Let $D  = \{x_1,\cdots,x_n\} \subseteq T$, where $T$ is a compact set over $\mbR^{d}$, be i.i.d. samples from a distribution with density $f_0$. For any given ICLP covariance kernel $K$, we adopt the ICLP-QR by picking the density estimation kernel as $K^{\eta}$ with $\eta >1$. For a given $d\times d$ symmetric and positive definite bandwidth matrix $\mathbf{H}$, the kernel density estimator under the ICLP-QR strategy takes the form of
\begin{equation}\label{eqn: kde form}
    \hat{K}_{D} (x) = \frac{1}{n} \sum_{i=1}^{n} K_{\mathbf{H}}^{\eta} \left( x-x_{i} \right)
    = \frac{1}{n \sqrt{det(\mathbf{H})}} \sum_{i=1}^{n} K^{\eta} \left( \mathbf{H}^{-\frac{1}{2}} (x-x_{i}) \right).
\end{equation}
We now provide the global sensitivity and utility analysis of $\hat{K}_{D} (x)$ in the following theorem.

\begin{theorem}\label{thm:kde gs and utility}
Suppose $K^{\eta}(\cdot,\cdot)$ is point-wise bounded by a constant $M_{K}$, then the global sensitivity $\Delta$ of $\hat{K}_{D} (x)$ in (\ref{eqn: kde form}) satisfies
\begin{equation*}
    \Delta = \sup_{D\sim D'} \left\| \hat{K}_{D} - \hat{K}_{D'} \right\|_{1,K}
     \leq \frac{2M_K}{n \sqrt{det(\mathbf{H})} } \sqrt{tr(K^{\eta-1})}.
\end{equation*}
Furthermore, taking $\mathbf{H}$ to be a diagonal matrix with the same entry, i.e., $\mathbf{H} = h\mathbf{I}$, and assuming $f_{0}^{''}$ is absolutely continuous, $\int_{T}(f_{0}^{'''}(x))^{2}dx < \infty$ and $\int_{T}K^{\eta}(x)dx = 1$. Then
\begin{equation*}
    \E \int_{T} \left( \tilde{f}_{D}(x) - f_0(x) \right)^2 dx  \leq O\left( \frac{c_{1}}{n^2 h^{2d}} + h^4 + \frac{c_{2}}{n h^d} \right),
\end{equation*}
for some constants $c_{1}$ and $c_{2}$.
\end{theorem}
 \begin{remark}
If $h$ is taken to be $h\asymp n^{\frac{1}{4+d}}$, then $R = O(n^{-\frac{4}{4+d}})$, which matches the optimal kernel density estimation rate \citep{wasserman2006all}.  
\end{remark}
The connection between estimating kernel and the noise kernel also appeared in \citet{hall2013differential}, where they stated that one could achieve $(\epsilon,\delta)$-DP by adding a Gaussian process with its covariance function equal to the kernel used in estimation. For privacy-safe bandwidth, $h$, we can pick $h\asymp n^{\frac{1}{4+d}}$ to ensure privacy is gained for free. However, a private version of the ``rule of thumb", see \citet{rao1992simple} and \citet{hall2013differential} is also feasible.

\paragraph{Functionals via Regularized ERM.}
The functional summaries one desires to release may come from learning algorithms such as regularization-based algorithms. In Section~\ref{ERM approach section}, we proposed using such algorithms to obtain qualified functional summaries. Here, we generalize the approach to broader scenarios such as non-parametric regression and classification. Let $\mathcal{D} = \{d_1,\cdots,d_n\}$ be the collection of $n$ samples, where $d_i$ is a tuple with finite size. Given a loss function $L$, we consider the following regularized ERM problem:
\begin{equation}\label{empirical risk formular}
    \hat{f}_D = \underset{f \in \mcH_{C^{\eta}}}{\operatorname{argmin}} \left\{ \frac{1}{n}\sum_{i=1}^{n} L(d_i,f) + \psi \|f\|_{C^{\eta}}  \right\} \quad \text{for some} \quad \eta > 1.
\end{equation}
When $d_i$'s are couples, i.e., $d_i = (y_i,x_i)$, (\ref{empirical risk formular}) can be viewed as non-parametric classification (where $y_i$'s take discrete value) or regression (where $y_i$'s take continuous value) problems. The solution of (\ref{empirical risk formular}) can be expressed as $\hat{f}_D = \sum_{i=1}^{n}a_i C^{\eta}(\cdot,d_i)$ by the Representer Theorem \citep{kimeldorf1971some}. However, although the Representer Theorem provides an elegant solution for (\ref{empirical risk formular}), it is not suitable for calculating the global sensitivity since all the elements in the vector $(a_1,\cdots,a_n)$ change when we swap one individual in the dataset. In the following theorem, we provide a sensitivity analysis for $\hat{f}_{D}$ under certain regularized conditions.
\begin{theorem}\label{thm:GS for ERM}
Suppose $\hat{f}_{D}$ is the solution of (\ref{empirical risk formular}) and the loss function $L$ in (\ref{empirical risk formular}) is an $M$-admissible loss function \citep{bousquet2002stability}, then the global sensitivity for $\hat{f}_{D}$ satisfies
\begin{equation*}
    \Delta = \sup_{D\sim D'}\left\| \hat{f}_{D} - \hat{f}_{D'} \right\|_{1,C} \leq \frac{M}{\psi n} \sqrt{\sup_{x}C^{\eta}(x,x)
     }\sqrt{tr(C^{\eta-1})}.
\end{equation*}
\end{theorem}
\begin{remark}
One can also prove that the privacy error $\E \|\tilde{f}_{D} - \hat{f}_{D}\|_{L^{2}}^{2}$ is bounded by $c_1 (\psi n)^{-2}$. We do not provide a utility analysis for this case study as the statistical error can vary based on different settings of the problem and is out of the scope of this paper.
\end{remark}
The application scenario is broad since the upper bound for the global sensitivity holds for any convex and locally $M$-admissible loss function and bounded kernel with finite trace. For example, support vector machines with hinge loss, non-parametric regressions with square loss, and logistic regressions with $log(1+x)$ loss are all applicable settings.

\subsection{Privacy-Safe Regularization Parameter Selection}\label{sec: privacy params}
Determining the regularization parameters in different mechanisms, such as the finite-dimensional number $M$ (for subspace embedding mechanisms) and $\psi$ (for the ICLP mechanism), is crucial to ensure the reasonable performance of the private releases. Tuning regularization parameters in statistical modeling has been well studied, and \textit{Cross Validation} (CV), or one of its many variants, is the most widely used approach. However, CV focuses on balancing variance and bias in the estimation error, not the trade-off between privacy and estimation error. To fit CV into the DP framework, \textit{Private Cross Validation} (PCV) was proposed in \citet{mirshani2019formal}, which aims to find the ``sweet spot" between privacy error and estimation error. However, as data-driven approaches, neither CV nor PCV is truly privacy-safe since the regularization parameters may contain information about the data. There are some approaches one can use to obtain end-to-end privacy-guaranteed regularization parameters. For example, one can use out-of-sample public datasets \citep{zhang2012functional} or one can spend extra privacy budget on the tuning process \citep{chaudhuri2011differentially,chaudhuri2013stability}.

Since the ICLP mechanism is tied to a kernel, one can obtain privacy-safe regularization parameters by picking kernels whose eigenvalues decay polynomially, satisfying the conditions in Theorem \ref{thm: total error for mean protection} so that the optimal values for $M$ and $\psi$ in Theorem \ref{thm: total error for mean protection} can be directly used as regularization parameter inputs. We name this approach \textit{``Privacy Safe Selection'' (PSS)}. PSS does not degrade the privacy guarantee since the employed regularization parameters rely only on the sample size, the privacy budget, and the additive noise's covariance function. We would like to note that PSS is not a data-driven approach since its acquisition never depends on the data. In practice, the constants for the optimal values in PSS can affect the performance of the ICLP mechanism. In our experiments, we observe that by appropriately normalizing the sample trajectories and the trace of the covariance kernel, setting the constant to $1$ usually leads to satisfactory performance.

\section{Algorithm and Implementation}\label{sec: algo and implement}

Based on the definition of the ICLP, the generic implementation of the mechanism can be achieved using the Karhunen-Lo\'eve expansion.
\begin{enumerate}
    \item Given any Mercer kernel $C$, obtain its eigenvalues $\{\lambda_{j}\}_{j\geq 1}$ and eigenfunctions $\{\phi_{j}\}_{j\geq 1}$.
    \item Generate ICLP noise by $Z = \sum_{j=1}^{\infty} \sqrt{\frac{\lambda_{j}}{2}} Z_j \phi_{j}$ where $Z_j \overset{i.i.d.}{\sim} \Lap(1)$.
    \item Calibrate $Z$ to desired privacy level by the global sensitivity $\Delta$ and privacy budget $\epsilon$.
\end{enumerate}

However, the summation in generating $Z$ cannot be implemented in finite time and usually is terminated at a large integer. Therefore, in practice, we utilize its approximated version, Algorithm \ref{approximated algorithm}.
\begin{algorithm}[ht]
\caption{Approximated ICLP mechanism}
\label{approximated algorithm}

Given the covariance kernel $C$ and $K$ different points $(x_{1},x_{2}, \cdots,x_{K})$ on the compact domain $\mcT$, calculate the value of $C$ on the grid expanded by $(x_{1},x_{2}, \cdots,x_{K})$, i.e., 
\begin{equation*}
    \hat{C} = 
    \begin{pmatrix}
    C(x_{1}, x_{1}) &  \cdots & C(x_{1}, x_{K}) \\
    \vdots   & \ddots & \vdots  \\
    C(x_{K}, x_{1}) & \cdots & C(x_{K}, x_{K}) 
    \end{pmatrix}
\end{equation*}

Obtain $K$ estimated eigenvalues $\{ \hat{\lambda}_{k} \}_{k=1}^{K}$ and eigenfunctions $\{ \hat{\phi}_{k} \}_{k=1}^{K}$ of $\hat{C}$ by eigendecomposition.

\For{k in 1,2,$\cdots$, K}
{
Set $\hat{f}_{Dk} =  \langle \hat{f}_D, \hat{\phi}_{k}  \rangle$ and generate $Z_k$ from $\sqrt{\frac{ \hat{\lambda}_k } {2}} \Lap(1)$
 \\
$\tilde{f}_{Dk} = \hat{f}_{Dk} + \sigma Z_k$ where $\sigma = \frac{\sqrt{2}\Delta}{\epsilon}$
}

Return $\tilde{f} = \sum_{k=1}^{K} \tilde{f}_{Dk} \phi_j$.
\end{algorithm}

A natural question about Algorithm \ref{approximated algorithm} is whether all the theoretical analyses still hold if the privacy noise is sampled in a finite approximation manner instead of the ``true'' infinite sum. The answer is positive as long as the same cutoff $K$ is used both in constructing privacy noise and expressing the original estimate, followed directly by the post-processing inequality. However, a key advantage of our theoretical analyses is that the privacy guarantees will still hold regardless of what $K$ is used. Another problem regarding Algorithm \ref{approximated algorithm} is that even though larger $K$ will lead to more accurate estimates of eigenvalues and eigenfunctions, it also increases the computational burden as the algorithm relies on the Karhunen-Lo\'eve expansion. Next, we investigated how different cutoff values, $K$, will affect computational time by comparing the average computation time for generating $100$ ICLPs to $100$ Gaussian Processes. We choose the Gaussian Process as the competitor since it is the stochastic process used to achieve $(\epsilon,\delta)$-DP for functional data, and sampling Gaussian processes is nothing more than sampling a multivariate Gaussian with covariance $\hat{C}$. Theoretically,  generating one ICLP and one Gaussian process are both in time complexity $O(n^{3})$ since both Cholesky decomposition and eigen decomposition are $O(n^3)$. In Table \ref{generation time comparison between ICLP and GP}, we report the average time to generate $100$ ICLPs and $100$ Gaussian Processes for different covariance kernels. We found that generating $100$ Gaussian Processes is about $30\%$ to $50\%$ faster than generating $100$ ICLPs in practice.

\begin{table}[ht]
\centering

\resizebox{\textwidth}{!}
{\begin{tabular}{|c|c|c|c||c|c|c|c|}
    \hline
    $C$ & $K$ & ICLP & GP & $C$ & $K$ & ICLP & GP \\
    \hline
    \multirow{3}{*}{Exponential} &  100 & 0.567688 & 0.273463 & \multirow{3}{*}{Mat\'ern($\nu =  \frac{3}{2}$)} &  100 & 0.571014 & 0.270682 \\
    &  200 & 2.778597 & 1.683808 & & 200 & 2.636642 & 1.617183 \\
    &  500 & 29.98454 & 20.90210 & & 500 & 29.38142 & 20.56870\\
    \hline
    \multirow{3}{*}{Gaussian} & 100 & 0.553025 & 0.268379 & \multirow{3}{*}{Mat\'ern($\nu =  \frac{5}{2}$)} &  100 & 0.551222 & 0.271477 \\
    &  200 & 2.617800 & 1.610193 & &  200 & 2.618398 & 1.615180 \\
    &  500 & 29.26284 & 20.41770 &  &  500 & 29.29077 & 20.53389 \\
    \hline
 \end{tabular} }

\captionsetup{width=\textwidth}
\caption{Computation time (in seconds) for generating $100$ ICLPs and Gaussian Processes under different cutoff values $K$ and covariance kernels $C$ over $[0,1]$.}
\label{generation time comparison between ICLP and GP}
\end{table}

\section{Experiments}\label{sec: simulation section}
In this section, we numerically evaluate the effectiveness of the ICLP mechanism and other comparable mechanisms, like the FRL and Bernstein mechanisms. 

\subsection{Simulation for Mean Function Protection}
In this section, we conduct the simulation for the mean function privacy protection problem discussed in Section \ref{sec: mean function protection}. We use the isotropic Mat\'ern kernel \citep{cressie1999classes} as the covariance kernel for the ICLP noise. It takes the form 
\begin{equation*}
    C_{\alpha}(s, t)=  \frac{1}{\Gamma(\nu) 2^{\alpha - 1} } \left(  \frac{\sqrt{2\alpha} d(s,t)}{\rho} \right)^{\alpha} K_{\alpha}\left( \frac{\sqrt{2\alpha} d(s,t)}{\rho}  \right)
\end{equation*}
where $K_{\alpha}$ is the modified Bessel function. This is motivated by the fact that the resulting RKHS of $C_{\alpha}$ ties to a particular Sobolev space, allowing us to control the smoothness directly. Specifically, the RKHS associated with the Mat\'ern kernel $C_{\alpha}$ over $\mathbb{R}^{d}$ is norm-equivalent to the Sobolev space $H^{\alpha + \frac{d}{2}}$ and thus the corresponding eigenvalues decay polynomially as $\lambda_{j}\asymp j^{-2(\alpha + \frac{d}{2})}$. In the following experiments, we set $d=1$, $\rho = 0.1$, the privacy budget $\epsilon = 1$, and $\alpha = 1.5$ such that $\lambda_{j}\asymp j^{-4}$. For the functional samples, we consider $\mbH = L^{2}([0,1])$ and generate samples as 
\begin{equation*}
    X_{i}(t) = \mu_{0}(t) + e_{i}(t)
\end{equation*}
where $e_{i}(t)$ are Gaussian processes with zero mean and radial basis function kernel as covariance function. For the true mean function $\mu_{0}$, we consider the following four different forms:
\begin{enumerate}[label=S-\arabic*: , wide=2em,  leftmargin=*]
    \item $\mu_{0}(t) = 10t * \exp(-t)$.
    \item $\mu_{0}(t) = 0.3 f_{0.3, 0.05}(t) + 0.7*f_{0.8, 0.05}(t)$.
    \item $\mu_{0}(t) = 0.2 * \left( f_{0, 0.03}(t) + f_{0.2, 0.05}(t) + f_{0.5, 0.05}(t) - f_{0.75, 0.03}(t) + f_{1, 0.03}(t) \right)$.
    \item $\mu_{0}(t) = \sum_{j=1}^{25} R_{ij} \phi_{j}(t)$, where $R_{ij} \overset{i.i.d.}{\sim} U[-1,1]$.
\end{enumerate}
Here, $f_{a,b}$ is the probability density function of the normal distribution with mean $a$ and variance $b^{2}$. The trajectory complexity of the samples generated from these mean functions rises sequentially. For example, S-1 is a monotonically increasing function, S-2 is a bimodal function, and S-3 and S-4 are functions exhibiting multiple rapid fluctuations. We evaluate these mechanisms via privacy-estimation errors, estimation errors, and privacy errors. These errors are calculated via Monte Carlo by generating $1000$ privatized mean estimators.

We also conduct the experiments in different settings. For example, we consider generating the error stochastic process $e_{i}$ from Gaussian processes with different covariance functions or from basis expansion with heavy-tailed distributions. We also consider the ICLP covariance kernel as the Mat\'ern kernel with $\alpha = 2.5$. We refer to Appendix~\ref{apd: additional experiments for mean function} for these additional experimental results.

\subsubsection{Comparison of PCV and PSS}\label{sec: compare PCV and PSS}

In Section \ref{sec: privacy params}, we introduced PSS for parameter selection. Here, we demonstrate its effectiveness by comparing it with the data-driven method PCV. For both the ICLP-AR and ICLP-QR, we set $\eta$ and $\psi_{\text{PSS}}$ to be the values in Theorem \ref{thm: total error for mean protection} such that the privacy error is the same order as the estimation error. For PCV, we obtain $\psi_{\text{PCV}}$ by $10$-fold PCV within the range of $[0.1 \psi_{\text{PSS}}, 10\psi_{\text{PSS}} ]$. In the FRL mechanism, the PSS approach is more ambiguous as the PSS values for truncation number such that $\tilde{\mu}$ reaches the optimal rate is a collection of integers, i.e., $\mathcal{M}= \{M_1,\cdots, M_K\}$. We calculate the estimation error for each $M\in \mathcal{M}$ and take the smallest estimation error as the PSS result. For PCV, we consider a wider range of $\mathcal{M}$ by adding and subtracting $3$ to its maximum and minimum elements. 

\begin{figure}[ht]
    \centering
    \includegraphics[page=1,width=1\linewidth]{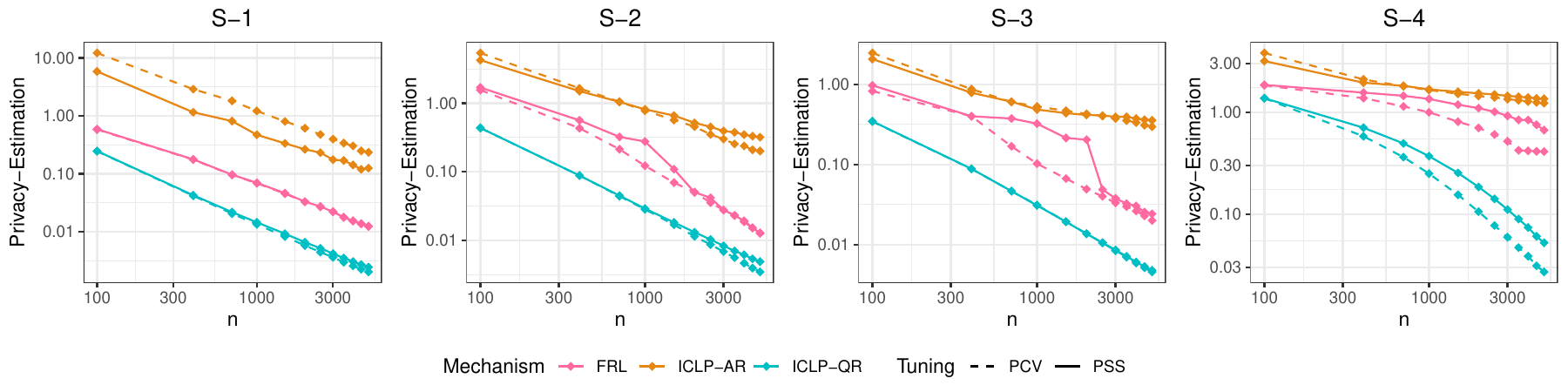} 
    \caption{Privacy-estimation error for PSS and PCV approaches to select parameter $\psi$ under different mechanisms, sample size, and true mean functions. Reported values are averages of $100$ independent replicated experiments. Both axes are in the base 10 log scale.
    }
    \label{fig: PCV PSS compare}
\end{figure}

In Figure \ref{fig: PCV PSS compare}, we report the privacy-estimation error for each mechanism under different sample sizes $n$. For the relatively simple $\mu_{0}$ in S-1, the error decay curves of the PSS almost line up with the PCV ones for FRL and ICLP-QR. In S-2 and S-3, the ICLP mechanism still has consistent error decay curves between the PSS and PCV, while the PSS curve of the FRL mechanism exhibits a step-down pattern. This pattern occurs because the maximum number in $\mathcal{M}$ is determined by the sample size. Therefore, with complex curves and small sample sizes, the FRL mechanism does not have enough components to estimate the mean function well. As the sample size increases, the availability of more components improves the estimation. In S-4, PCV performs slightly better than the PSS approach for both mechanisms. This is expected since the FRL mechanism requires more dimensions, ICLP-QR requires less regularization, and PCV consistently behaves this way. Since it has been shown that selecting regularization parameters via the PSS approach provides a reasonable and consistent performance compared to PCV, we use the PSS approach in the following experiments to be fully privacy-safe.

\subsubsection{Comparison of Different Mechanisms}

Under the same settings, we compare the performance of different mechanisms under different sample sizes $n$, which include the FRL, ICLP-AR, ICLP-QR, and the Bernstein mechanisms\footnote{The implementation of the Bernstein mechanism is based on R package \texttt{diffpriv}. We use the sample mean $\bar{X}$ as the non-private summary by setting the cover size parameter as $20$.}. We also include the Gaussian mechanism for achieving $(\epsilon,\delta)$-DP (with $\epsilon = 1$ and $\delta = 0.01$) on functional summaries via Gaussian process \citep{mirshani2019formal}, and we refer to it as GP-ADP. This provides insight into what is gained by moving from $\epsilon$-DP to $(\epsilon,\delta)$-DP.  
\begin{figure}[ht]
    \centering
        \includegraphics[width = \textwidth]{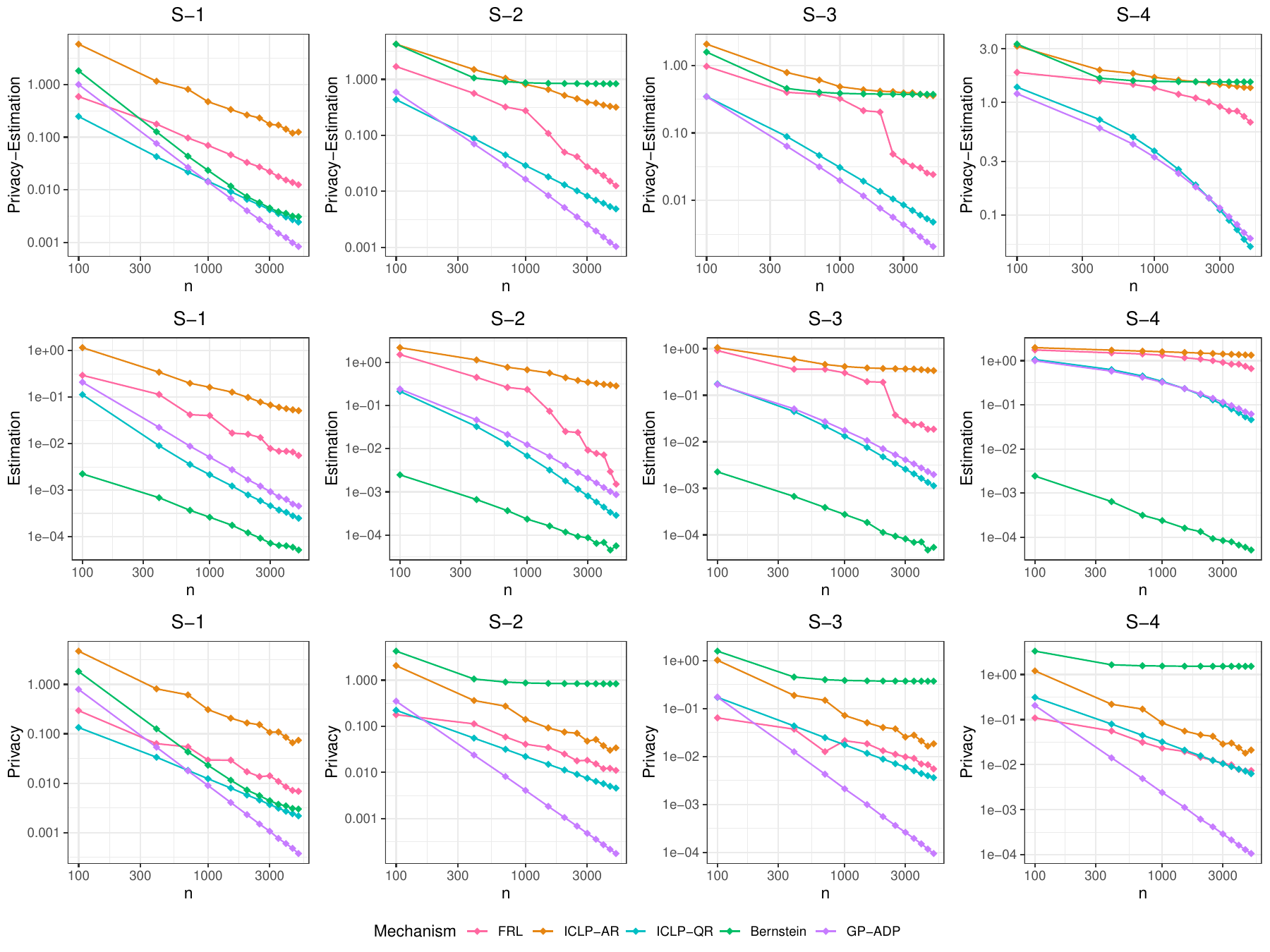}
    \caption{Privacy-estimation, estimation, and privacy errors for different mechanisms under different sample sizes $n$ and true mean functions. The values reported are averages over $100$ independent replicated experiments. Both axes are in the base 10 log scale.}
    \label{fig: nu32 pss}
\end{figure}

Figure~\ref{fig: nu32 pss} illustrates the error decay results of different mechanisms as $\mu_{0}$ and $n$ are varied. Focusing on the privacy-estimation error, the ICLP-QR consistently outperforms all other $\epsilon$-DP mechanisms across various scenarios. This indicates that the ICLP-QR is effective in releasing privatized summaries with high utility. On the other hand, GP-ADP performs slightly better than the ICLP-QR, especially when $n$ is larger. This is not surprising, as moving to a more relaxed privacy paradigm can lead to private summaries with higher utility. On the other hand, the ICLP-AR exhibits the worst performance, primarily due to its high estimation error. This poor performance is likely caused by the exact bias introduced by the soft thresholding function, which would require an extremely large sample size to reduce the threshold and eliminate the bias. The FRL and Bernstein mechanisms are close to the ICLP-QR in S-1, showing their effectiveness in simpler mean function scenarios, but they fail to mimic the behaviors of the ICLP-QR when mean functions become more complex, i.e., S-2 to S-4. 

Regarding the trade-off between estimation and privacy error, the ICLP-QR mechanism exhibits similar privacy error to the FLR mechanism while significantly outperforming it in estimation error. This confirms the ICLP mechanism's advantages from two perspectives. First, by treating the summary as infinite-dimensional, the ICLP mechanism provides better non-private estimations than the finite-dimensional FLR mechanism, which suffers from worse estimation errors due to fewer components involved. Second, despite adding noise to infinite dimensions, the ICLP mechanism only requires a similar amount of noise as the FLR mechanism, indicating it achieves more effective noise injection by treating different dimensions heterogeneously.

\subsection{Simulation for Kernel Density Estimator Protection}
To demonstrate the wide range of application scenarios of the ICLP mechanism, we conduct simulations on kernel density estimations. We consider the setting under $\mbR$ and $\mbR^2$ with samples generated from two mixture Gaussian distributions.

\begin{figure}[ht]
     \centering
     \begin{subfigure}[b]{0.25\textwidth}
         \centering
         \includegraphics[page=1,width=\textwidth]{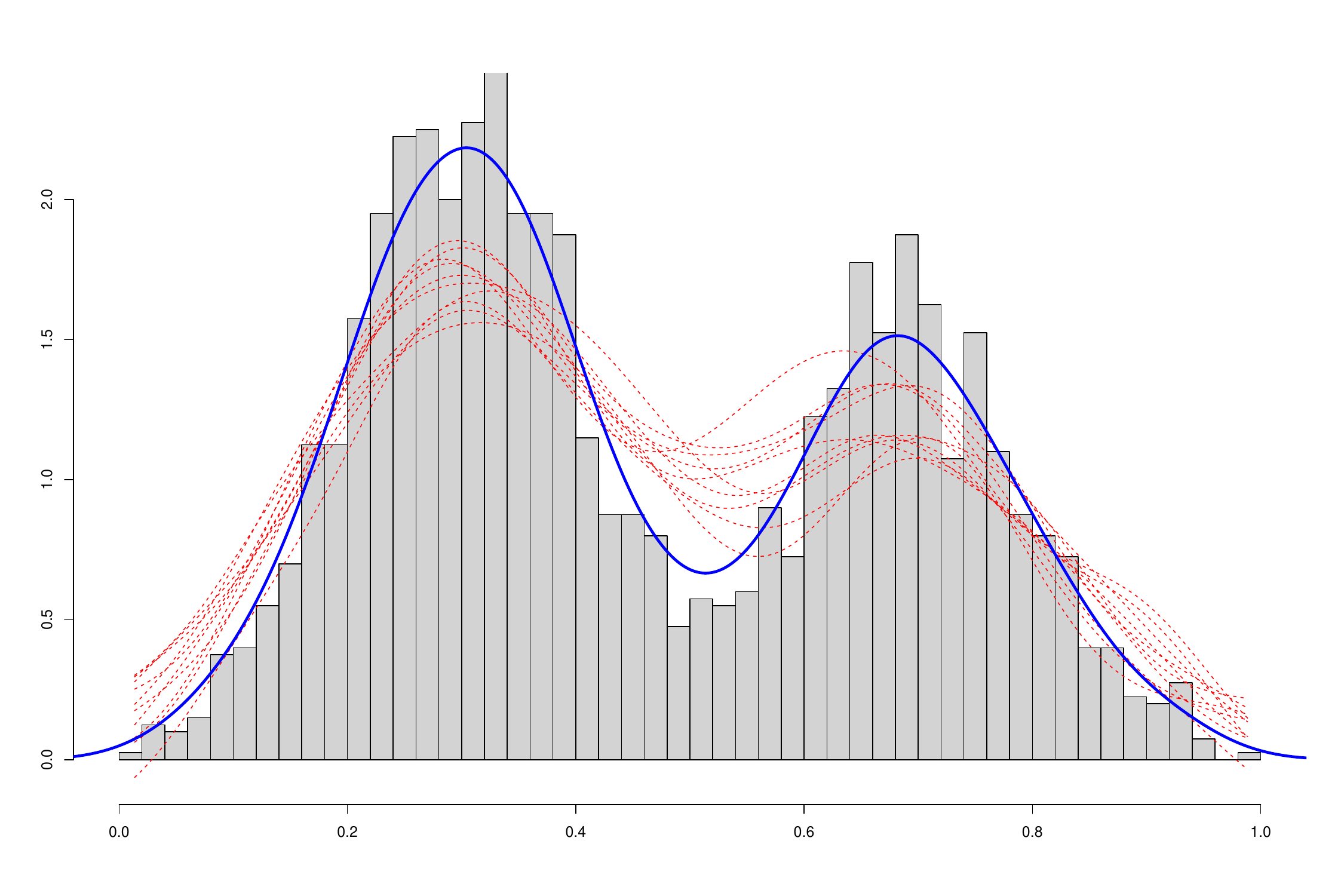}
        \caption{}
     \end{subfigure}
     \hspace{1em}
     \begin{subfigure}[b]{0.25\textwidth}
         \centering
         \includegraphics[page=2,width=\textwidth]{Figures/Simulation/density/1d_plot/1d_density_comp_gau.pdf}
          \caption{}
     \end{subfigure}
     \hspace{1em}
     \begin{subfigure}[b]{0.25\textwidth}
         \centering
         \includegraphics[page=3,width=\textwidth]{Figures/Simulation/density/1d_plot/1d_density_comp_gau.pdf}
         \caption{}
     \end{subfigure}
     \vskip\baselineskip
    \begin{subfigure}[b]{0.25\textwidth}
         \centering
         \includegraphics[page=4,width=\textwidth]{Figures/Simulation/density/1d_plot/1d_density_comp_gau.pdf}
        \caption{}
     \end{subfigure}
     \hspace{1em}
     \begin{subfigure}[b]{0.25\textwidth}
         \centering
         \includegraphics[page=5,width=\textwidth]{Figures/Simulation/density/1d_plot/1d_density_comp_gau.pdf}
          \caption{}
     \end{subfigure}
     \hspace{1em}
     \begin{subfigure}[b]{0.25\textwidth}
         \centering
         \includegraphics[page=6,width=\textwidth]{Figures/Simulation/density/1d_plot/1d_density_comp_gau.pdf}
         \caption{}
     \end{subfigure}
     
    \caption{Non-private \textcolor{blue}{(Blue)} and 10 random realization of private \textcolor{red}{(Red)} KDEs. (a) ICLP mechanism with $\eta = 1.25$ (b) ICLP mechanism with $\eta = 1.5$ (c) FRL and (d)-(f) Bernstein mechanism with lattice number equal to $10,30,50$.}
    \label{fig:1d density plots}
\end{figure}

\begin{figure}[ht]
     \centering
     \begin{subfigure}[b]{0.25\textwidth}
         \centering
         \includegraphics[page=1,width=\textwidth]{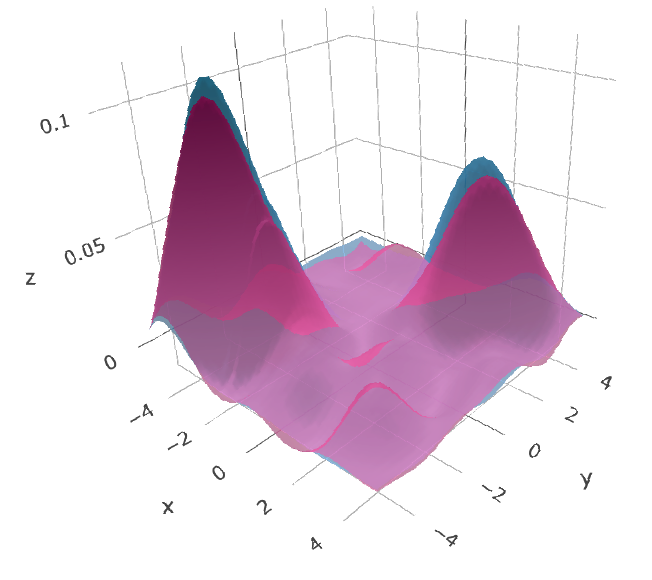}
        \caption{}
        \label{2d iclp 101}
     \end{subfigure}
     \hspace{1em}
     \begin{subfigure}[b]{0.25\textwidth}
         \centering
         \includegraphics[page=2,width=\textwidth]{Figures/Simulation/density/2d_plot/2d_exp_2000.pdf}
          \caption{}
        \label{2d iclp 105}
     \end{subfigure}
     \hspace{1em}
     \begin{subfigure}[b]{0.25\textwidth}
         \centering
         \includegraphics[page=3,width=\textwidth]{Figures/Simulation/density/2d_plot/2d_exp_2000.pdf}
         \caption{}
        \label{2d iclp 12}
     \end{subfigure}
     \vskip\baselineskip
    \begin{subfigure}[b]{0.25\textwidth}
         \centering
         \includegraphics[page=4,width=\textwidth]{Figures/Simulation/density/2d_plot/2d_exp_2000.pdf}
        \caption{}
     \end{subfigure}
     \hspace{1em}
     \begin{subfigure}[b]{0.25\textwidth}
         \centering
         \includegraphics[page=5,width=\textwidth]{Figures/Simulation/density/2d_plot/2d_exp_2000.pdf}
          \caption{}
     \end{subfigure}
     \hspace{1em}
     \begin{subfigure}[b]{0.25\textwidth}
         \centering
         \includegraphics[page=6,width=\textwidth]{Figures/Simulation/density/2d_plot/2d_exp_2000.pdf}
         \caption{}
     \end{subfigure}
     
    \caption{ 3D plot of non-private \textcolor{blue}{(Blue)} and private \textcolor{red}{(Red)} KDEs over $\mbR^{2}$. (a)-(c) ICLP mechanism with $\eta = 1.01, 1.05, 1.2$  (d) FRL and (e)-(f) Bernstein mechanism with lattice number equal to $5,10$.}
    \label{fig:2d_density_exp}
\end{figure}

\begin{enumerate}
    \item $\mbR$ setting:
    \begin{equation*}
        x_i \stackrel{i.i.d.}{\sim} \sum_{i=1}^{2}p_{i}\mathcal{N}(\mu_i, 0.1; 0,1),
    \end{equation*}
    where $\mathcal{N}(\mu,\sigma; a, b)$ is a truncated normal distribution over $[a,b]$ with $p_1 = 0.6$, $p_2 = 0.4$, $\mu_1 = 0.3$, $\mu_2 = 0.7$.
    
    \item $\mbR^2$ setting:
    \begin{equation*}
        \mathbf{x}_i \stackrel{i.i.d.}{\sim} \sum_{i=1}^{2} p_{i}\mathcal{N}\left( \mu_{i}, \big(\begin{smallmatrix}
    1 & 0.5\\
    0.5 & 1
    \end{smallmatrix}\big); \big(\begin{smallmatrix}
    5\\
    -5
    \end{smallmatrix}\big),\big(\begin{smallmatrix}
    5\\
    -5
    \end{smallmatrix}\big)\right) ,
    \end{equation*}
    where $\mathcal{N}(\mu,\Sigma; \mathbf{a},\mathbf{b})$ is a multivariate truncated normal distribution over $[a_1,b_1]\times [a_2,b_2]$ with $p_1 = 0.6$, $p_2 = 0.4$, $\mu_1 = (-3,-3)$, $\mu_2 = (3,2)$.
\end{enumerate}
We compare the FRL, ICLP-QR, and Bernstein mechanisms. For the ICLP-QR and the Bernstein mechanism, we pick multiple smoothing parameters $\eta$ and lattice numbers to demonstrate how they affect private curves and surfaces. For the FRL mechanism, we select the truncated number that provides the best fit under PCV criteria. We use the Gaussian kernel in $\mbR$ and the exponential kernel in $\mbR^2$ to build the kernel density estimator. We use $h\asymp n^{1/(4+d)}$ where $d=1,2$ to ensure we gain privacy for free and remain privacy safe. The results are reported in Figure \ref{fig:1d density plots} and Figure \ref{fig:2d_density_exp}.

For the univariate setting, the ICLP-QR performs similarly to the FRL mechanism; a higher $\eta$ produces less variability in the curves but tends to be over-smoothed. The Bernstein mechanism needs over $30$ lattice points in the interval to capture the bimodal pattern but results in producing a messy tail at both ends. A lower lattice number produces better tails but fails to capture the bimodal pattern.

For the multivariate setting, one can see that by slightly oversmoothing, the ICLP-QR produces privatized KDEs that are very close to the non-private ones. A smaller $\eta$ (Figure \ref{2d iclp 101}) is more precise at peaks but will be ``noisy'' around lower density regions, while a larger $\eta$ (Figure \ref{2d iclp 12}) produces smooth lower density regions but causes underestimation at peaks. Figure \ref{2d iclp 105} shows there is a clear ``sweet point'' to trade off the smoothness and underestimation. The FRL mechanism performs similarly to the underestimated ICLP case, but the peaks of the privatized KDE do not fully align with the non-private one. On the other hand, the Bernstein mechanism fails to produce surfaces similar to those of the non-private estimator, even when we increase the number of lattice points.

\section{Real Data Applications}\label{sec: real data section}
This section presents two real data applications of the proposed methods to study the release of functional summaries for different functional data datasets.

\subsection{Application on Medical and Energy Usage Functional Data}
This application aims to release a private mean function that satisfies $\epsilon$-DP for the following two functional data datasets. The first dataset is the Brain scans Diffusion Tensor Imaging (DTI) dataset\footnote{Available in the R package \texttt{refund}.}. The DTI dataset provides fractional anisotropy (FA) tract profiles for the corpus callosum (CCA) of the right corticospinal tract (RCST) for patients with multiple sclerosis and for controls. Specifically, we study the CCA dataset, which includes 382 patients measured at 93 equally spaced locations of the CCA. The second dataset contains historical electricity demand in Adelaide\footnote{Available in the R package \texttt{fds}.}. The dataset consists of half-hourly electricity demands from Sunday to Saturday in Adelaide between July 6, 1997, and March 31, 2007. Our analysis focuses on Monday specifically, meaning the dataset consists of measurements from 508 days at 48 equally spaced time points. 

\begin{table}[ht]
\centering
\resizebox{\columnwidth}{!}
{\begin{tabular}{|r|cccc|cccc|}
\hline
& \multicolumn{8}{c|}{Eletricity Demand} \\ \cline{2-9}
& \multicolumn{4}{c|}{ $C_{1.5}$ } & \multicolumn{4}{c|}{ $C_{2.5}$ }\\
\hline
$\epsilon$ &  FRL & ICLP-AR & ICLP-QR & Bernstein &  FRL & ICLP-AR & ICLP-QR & Bernstein\\
\hdashline
  1/8 & $0.2828_{0.1981}$ & $2.2110_{1.5374}$ & $3.9110_{2.9464}$ & $1.3726_{1.2935}$ & $0.2805_{0.1645}$ & $2.2279_{1.5880}$ & $4.1614_{3.5764}$ & $1.4261_{1.3547}$ \\ 
  1/4 & $0.1731_{0.0419}$ & $0.6140_{0.3859}$ & $1.0391_{0.8066}$ & $0.3459_{0.2783}$ & $0.1703_{0.0376}$ & $0.6053_{0.4227}$ & $1.1168_{0.8931}$ & $0.3468_{0.2963}$ \\ 
  1/2 & $0.1453_{0.0098}$ & $0.2117_{0.1276}$ & $0.3415_{0.1897}$ & $0.0895_{0.0686}$ & $0.1427_{0.0094}$ & $0.2092_{0.0954}$ & $0.3497_{0.2276}$ & $0.0920_{0.0692}$ \\ 
  1 & $0.1383_{0.0025}$ & $0.1106_{0.0352}$ & $0.1649_{0.0728}$ & $0.0241_{0.0171}$ & $0.1360_{0.0026}$ & $0.1109_{0.0374}$ & $0.1639_{0.0655}$ & $0.0251_{0.0168}$ \\ 
  2 & $0.1366_{0.0007}$ & $0.0858_{0.0151}$ & $0.1204_{0.0238}$ & $0.0080_{0.0044}$ & $0.1342_{0.0006}$ & $0.0857_{0.0167}$ & $0.1164_{0.0233}$ & $0.0089_{0.0049}$ \\ 
  4 & $0.1362_{0.0002}$ & $0.0793_{0.0074}$ & $0.1089_{0.0105}$ & $0.0039_{0.0011}$ & $0.1338_{0.0002}$ & $0.0794_{0.0072}$ & $0.1044_{0.0090}$ & $0.0048_{0.0012}$ \\ 
    \hhline{=========} 
  & \multicolumn{8}{c|}{ DTI }  \\ \cline{2-9}
  & \multicolumn{4}{c|}{ $C_{1.5}$ } & \multicolumn{4}{c|}{ $C_{2.5}$ }\\
  \hline
  $\epsilon$ &  FRL & ICLP-AR & ICLP-QR & Bernstein &  FRL & ICLP-AR & ICLP-QR & Bernstein\\
\hdashline
  2 & $0.6305_{0.2878}$ & $6.2473_{4.2907}$ & $4.4245_{3.9617}$ & $3.1996_{2.6794}$ & $0.6303_{0.2867}$ & $6.2155_{4.8556}$ & $4.7453_{4.0892}$ & $3.0841_{2.7095}$ \\ 
  3 & $0.4418_{0.0663}$ & $1.6122_{1.0886}$ & $1.2691_{1.1059}$ & $0.8039_{0.6994}$ & $0.4437_{0.0762}$ & $1.6275_{1.1514}$ & $1.3049_{0.9318}$ & $0.7820_{0.7225}$ \\ 
  4 & $0.3952_{0.0165}$ & $0.4929_{0.3046}$ & $0.4701_{0.2522}$ & $0.2040_{0.1627}$ & $0.3974_{0.0158}$ & $0.4907_{0.3274}$ & $0.4793_{0.2431}$ & $0.2028_{0.1790}$ \\ 
  5 & $0.3836_{0.0045}$ & $0.2080_{0.0824}$ & $0.2747_{0.0756}$ & $0.0555_{0.0369}$ & $0.3859_{0.0040}$ & $0.2073_{0.0890}$ & $0.2748_{0.0804}$ & $0.0581_{0.0395}$ \\ 
  6 & $0.3809_{0.0010}$ & $0.1361_{0.0285}$ & $0.2265_{0.0265}$ & $0.0186_{0.0106}$ & $0.3830_{0.0010}$ & $0.1366_{0.0287}$ & $0.2231_{0.0325}$ & $0.0216_{0.0107}$ \\ 
  7 & $40.3801_{0.0003}$ & $0.1187_{0.0133}$ & $0.2143_{0.0133}$ & $0.0094_{0.0028}$ & $0.3823_{0.0003}$ & $0.1187_{0.0137}$ & $0.2102_{0.0126}$ & $0.0126_{0.0029}$ \\ 
  \hline
\end{tabular}}
\captionsetup{width=1\textwidth}
\caption{Expected $L^{2}$-distance between the private mean function and the sample mean for both electricity demand and DTI(cca) datasets. The numbers in the subscript indicate the standard error ($\times10^{-3}$).}
\label{table: relativeMSE for energy and DTI data}
\end{table}

\begin{figure}[ht]
    \centering
    \includegraphics[width = \textwidth]{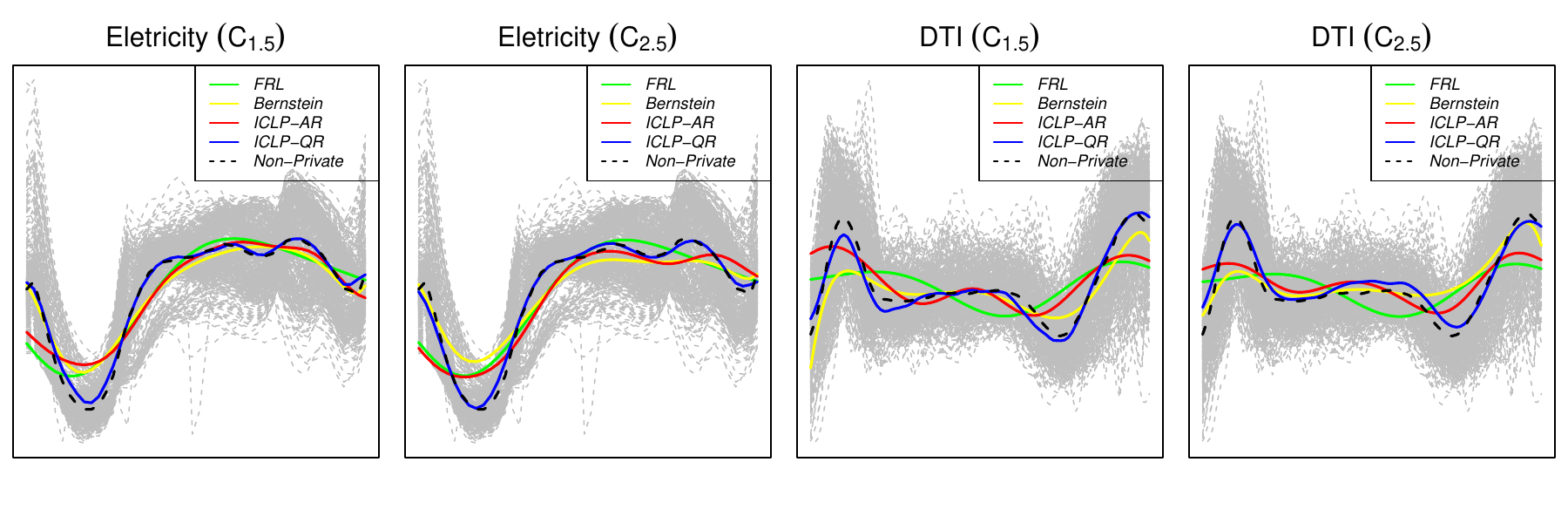}
    \caption{Non-private and private mean functions for different mechanisms with $\epsilon = 1$. The curves in light grey indicate the original samples.}
    \label{fig: real data private mean curves}
\end{figure}

Since the true mean function is not available in real data applications, we evaluate the performance of each mechanism using the expected $L^{2}$-distance between the private summary, $\tilde{\mu}$, and the non-private sample means, $\hat{\mu}$, i.e., $\E\|\tilde{\mu} - \hat{\mu}\|_{L^{2}}^{2}$. We consider the Mat\'ern kernel with $\alpha = 1.5$ and $2.5$ and $\rho = 0.1$. The expected $L^{2}$-distance is approximated using Monte Carlo with $1000$ generated $\tilde{\mu}$. The results are reported in Table \ref{table: relativeMSE for energy and DTI data}, with each value being an average of $100$ replicate experiments. We also plot one private mean estimator for each mechanism in Figure \ref{fig: real data private mean curves}. 

From Table \ref{table: relativeMSE for energy and DTI data}, it can be observed that the expected $L^{2}$-distance decreases similarly as the privacy budget increases for both datasets. One can see that the expected $L^{2}$-distance of the FRL soon stops changing, indicating that most of the errors of the FRL are concentrated on estimation errors. This indicates that in order to avoid adding too much noise to the later components, the FRL mechanism has to compromise on using fewer leading components, leading to higher estimation errors. This can also be seen in Figure~\ref{fig: real data private mean curves}, where the FRL mechanism only produces a privatized mean function that estimates the overall shape but fails to capture local details. The ICLP-AR and the Bernstein mechanisms have similar performance patterns and much worse results with small privacy budgets. Finally, the ICLP-QR performs the best among all approaches as its privatized mean functions can estimate the shapes precisely and have much smaller expected $L^{2}$-distances compared to the non-private mean.

\subsection{Application on Human Mortality Data}\label{subsec: age-at-death KDE}
Publishing the entire age-at-death distribution for a given country or region usually provides more comprehensive information about human lifespan and health status than publishing crude mortality rates. A private version of this distributional summary ensures that an attacker cannot infer information about individuals or groups in a particular age range. The mortality data for each region are collected from the United Nations World Population Prospects 2019 Databases\footnote{Available at \url{https://population.un.org/wpp/Download}.}. The dataset records the number of deaths for each region and age. The goal of this application is to release private mortality distributions for various regions.

\begin{table}[ht]
\centering
\resizebox{\columnwidth}{!}{
\begin{tabular}{c p{20mm}p{20mm}p{20mm}p{20mm}p{20mm}}
  \hline
     & \multicolumn{5}{c}{Mechanisms} \\ \cline{2-6}
    \multirow{2}{*}{Region} &   ICLP-QR \newline ($\eta=1.01$) & ICLP-QR \newline ($\eta=1.05$) & FRL & Bernstein \newline ($K=10$) & Bernstein \newline ($K=20$) \\ 
    \hline
  Eastern Africa & $\textbf{1.685}_{10.507}$ & $5.249_{7.549}$ & $5.252_{7.881}$ & $3.823_{12.243}$ & $2.400_{11.896}$ \\ 
  Middle Africa & $\textbf{1.122}_{0.451}$ & $5.281_{0.410}$ & $4.870_{2.152}$ & $3.969_{0.313}$ & $2.090_{0.327}$ \\ 
  Northern Africa & $\textbf{2.398}_{0.680}$ & $8.873_{0.656}$ & $3.991_{1.807}$ & $11.315_{0.546}$ & $5.710_{0.646}$ \\ 
  Southern Africa & $2.487_{1.491}$ & $6.427_{0.919}$ & $2.990_{1.809}$ & $3.964_{0.666}$ & $\textbf{2.395}_{0.763}$ \\ 
  Western Africa & $\textbf{1.588}_{13.699}$ & $5.775_{9.216}$ & $5.752_{11.482}$ & $4.347_{15.649}$ & $2.609_{15.493}$ \\ 
  Central Asia & $\textbf{5.804}_{2.715}$ & $11.829_{1.547}$ & $7.319_{3.739}$ & $13.163_{1.346}$ & $8.116_{1.808}$ \\ 
  Eastern Asia & $\textbf{3.551}_{4.414}$ & $13.446_{6.872}$ & $3.900_{6.958}$ & $15.878_{14.110}$ & $8.162_{9.175}$ \\ 
  Southern Asia & $\textbf{2.040}_{2.267}$ & $8.619_{3.737}$ & $2.968_{2.962}$ & $10.044_{8.099}$ & $4.962_{4.888}$ \\ 
  South-Eastern Asia & $\textbf{2.259}_{1.871}$ & $9.498_{3.565}$ & $2.566_{2.323}$ & $10.428_{8.328}$ & $5.157_{4.778}$ \\ 
  Western Asia & $\textbf{2.195}_{0.585}$ & $8.689_{0.624}$ & $3.768_{2.005}$ & $9.380_{0.510}$ & $4.640_{0.578}$ \\ 
  Eastern Europe & $\textbf{3.990}_{3.658}$ & $13.501_{4.924}$ & $4.817_{5.762}$ & $15.542_{9.356}$ & $8.438_{6.414}$ \\ 
  Northern Europe & $7.318_{1.626}$ & $19.087_{1.187}$ & $\textbf{7.174}_{1.493}$ & $22.846_{1.218}$ & $13.339_{1.304}$ \\ 
  Southern Europe & $7.457_{0.977}$ & $21.076_{0.931}$ & $\textbf{7.137}_{1.083}$ & $27.500_{0.852}$ & $15.873_{0.927}$ \\ 
  Western Europe & $6.819_{2.184}$ & $19.959_{2.782}$ & $\textbf{6.445}_{3.417}$ & $25.606_{5.375}$ & $14.760_{3.965}$ \\ 
  Caribbean & $6.453_{2.890}$ & $11.247_{2.022}$ & $\textbf{5.981}_{3.634}$ & $10.714_{1.580}$ & $7.231_{2.296}$ \\ 
  Central America & $1.636_{0.548}$ & $7.777_{0.685}$ & $\textbf{1.146}_{0.537}$ & $5.637_{0.480}$ & $2.616_{0.448}$ \\ 
  South America & $\textbf{2.244}_{2.498}$ & $9.461_{4.096}$ & $2.886_{3.276}$ & $9.294_{6.744}$ & $4.700_{4.333}$ \\ 
  Northern America & $2.205_{1.734}$ & $10.605_{2.792}$ & $\textbf{1.537}_{2.297}$ & $10.028_{5.546}$ & $4.746_{3.153}$ \\ 
  Australia/New Zealand & $23.525_{11.299}$ & $29.495_{5.040}$ & $\textbf{22.782}_{8.189}$ & $28.339_{5.053}$ & $23.983_{7.582}$ \\ 
    \hline 
 \end{tabular}
 }
\captionsetup{width=1\textwidth}
\caption{Expected $L^{2}$-distance between release KDEs and non-private KDEs for each region with $\epsilon = 1$. The numbers in the subscript indicate the standard error ($\times10^{-3}$).}
\label{KDE L2 error table}
\end{table}

We estimate the probability density function for each region and privatize the estimates via the ICLP mechanism and its competitors. The privacy budget is set to $1$. We evaluate the performance using $\E\| \tilde{f} - \hat{f} \|_{L^{2}}^{2}$ where $\tilde{f}$ is the private KDE and $\hat{f}$ is the non-private counterpart. The expectation is approximated using Monte Carlo using $200$ private KDEs. The results are reported in Table \ref{KDE L2 error table}. We also visualized the private KDEs for each region and mechanism in Appendix~\ref{apd: Visualization of Age-at-Death KDE}. 

From Table \ref{KDE L2 error table}, the ICLP-QR with $\eta = 1.01$ has smaller errors in developing regions, while the FRL mechanism performs better in developed regions. This is reasonable, as developed regions usually have better medical conditions, making the mortality age concentrate between $70$ and $80$ and their densities unimodal. Thus, a few leading components are sufficient to represent the density function in these regions. Conversely, the situation is different in developing regions, where the infant mortality rates are higher, making their densities multimodal and requiring more components for better estimation.

\section{Conclusion}\label{sec: conclusion section}
In this paper, we introduce a new mechanism, the ICLP mechanism, to achieve $\epsilon$-DP for infinite-dimensional functional summaries. This mechanism offers a wide range of output privacy protections with more flexible data assumptions and a more effective noise injection process compared to current mechanisms that rely on finite-dimensional embedding. We establish its feasibility in separable Hilbert spaces and spaces of continuous functions. Different approaches are proposed for constructing qualified summaries compatible with the ICLP mechanism, along with parameter selection via PSS, to guarantee end-to-end protection. We also demonstrate that one can balance utility and privacy by controlling the degree of regularization in these strategies. This is demonstrated in the mean protection example by showing that slightly over-smoothing the summary allows the private summary to achieve the optimal rate for non-private mean estimation.

Despite its advantages, the ICLP mechanism has some limitations and presents opportunities for future research. As discussed in Section \ref{sec: algo and implement}, implementing the ICLP mechanism relies on the Karhunen-Lo\'eve expansion, which can be computationally expensive. Developing a computational approach that does not depend on the Karhunen-Lo\'eve expansion is an important future direction. Additionally, although various experimental results indicate that by appropriately processing the sample trajectories and the ICLP covariance kernel, omitting the constant for PSS values can produce satisfactory performance, we believe a more careful investigation of the constant can further enhance performance.


\acks{
This work was partially supported by the National Science Foundation, NSF SES-1853209.
}




\newpage
{\noindent \LARGE \textbf{Appendix}} \par 
\noindent 
\appendix

\section*{Table of Contents}
\startcontents[sections]
\printcontents[sections]{}{1}{}

\section{Proofs for Section \ref{sec: feasibility section}}\label{apd: proofs for ICLP}

\subsection{Proof of Theorem \ref{thm: existence of iclp}}
\begin{proof}
Let $X$ be a random element defined via decomposition (\ref{eqn: ICLP KL-decomposition}) with covariance operator $C$, where $C$ belongs to the trace class. To prove that $X$ is well-defined in $\mathbb{H}$, we only need to show $\E \langle X, X \rangle_{\mbH} < \infty$. Note that
\begin{equation*}
    \E \langle X, X \rangle_{\mbH}  = \E \sum_{j=1}^{\infty}\sum_{k\geq 1} \sqrt{\lambda_j}\sqrt{\lambda_k} Z_{j} Z_{k} \langle\phi_{j}, \phi_{k} \rangle_{\mbH}= \E \left[ \sum_{j=1}^{\infty} \lambda_j Z_{j}^{2} \right].
\end{equation*}
Since $Z_{j}$ are i.i.d. Laplace random variables with zero mean and variance $1$, we have 
\begin{equation*}
    \E \left[ \lambda_j Z_{j}^{2} \right] = \sum_{j=1}^{\infty}\lambda_j  = \text{Tr}(C)< \infty.
\end{equation*}
Then, by the Fubini theorem,
\begin{equation*}
    \E \langle X, X \rangle_{\mbH} = \E \left[ \sum_{j=1}^{\infty} \lambda_j Z_{j}^{2} \right]= \sum_{j=1}^{\infty} \E \left[ \lambda_j Z_{j}^{2} \right]  < \infty,
\end{equation*}
which proves the existence of $X$.
\end{proof}

\subsection{Proof of Theorem~\ref{Thm: Equivalence of ICLP probability measure}}

\begin{proof}
We first define an isometric isomorphism between the Hilbert space $\mbH$ and the $\ell^{2}$ space to avoid considering probability measures over $\mbH$. Given an orthonormal basis $\{\phi_j\}_{j=1}^{\infty}$ of $\mbH$, define a mapping $\mcT : \mathbb{H} \rightarrow \ell^{2}$ by $\mcT(f) = \{ \langle f,\phi_j \rangle \}_{j=1}^{\infty}$. The inverse of $\mcT$ is $\mcT^{-1}( \{ \langle f,\phi_j \rangle )_{j=1}^{\infty}\} = \sum_{j=1}^{\infty}\langle f,\phi_j \rangle \phi_j$ and $\mcT$ preserves the norms, i.e., $\|\mcT(f)\|_{\ell^2} = \|f\|_{\mathbb{H}}$. Thus, this mapping is an isometric isomorphism between $\mbH$ and $\ell^{2}$, and we can consider the probability measure over $\ell^{2}$ rather than over $\mathbb{H}$.

For a Laplace random variable $X \sim \Lap(\mu,b)$ over $\mathbb{R}$, it induces a probability measure over $(\mathbb{R},\mathcal{B})$, where $\mathcal{B}$ is the Borel set over $\mathbb{R}$, as
\begin{equation*}
    \gamma_{\mu,b}(dx) = \frac{1}{2b}\exp\left( -\frac{|x-\mu|}{b} \right) dx.
\end{equation*}
Denote $\{\lambda_{j}\}_{j\geq 1}$ and $\{\phi_{j}\}_{j\geq 1}$ as the eigenvalues and eigenfunctions of the covariance operator $C$, respectively, and let $\lambda_j = 2b_{j}^2$. By the Existence of Product Measures Theorem \citep{tao2011introduction}, $P_{D}\circ \mcT$ is a unique probability measure defined as $\gamma(f_{D}, C) := \prod_{j=1}^{\infty}\gamma_{f_{Dj},b_j}$ over $(\mathbb{R}^{\infty}, \mathcal{B}^{\infty}): = (\prod_{j=1}^{\infty} \mathbb{R}_j, \prod_{j=1}^{\infty} \mathcal{B}_j)$. We further restrict $\gamma(f_{D}, C)$ on $(\ell^{2}, \sigma(\ell^{2}))$ and keep denoting it by $\gamma(f_{D}, C)$. Since $\mcT$ is an isometric isomorphism mapping between $\mcH$ and $\ell^{2}$ space, to prove $P_{D}$ and $P_{D'}$ are equivalent, it is sufficient to prove that the product probability measures $\gamma(f_{D}, C)$ and $\gamma(f_{D'}, C)$ are equivalent. Now, we prove Theorem~\ref{Thm: Equivalence of ICLP probability measure} by showing $\gamma(h, C)$ and $\gamma(0, C)$ are equivalent if and only if $h\in\mcH_{C}$. 

For the ``if'' part, we apply Kakutani's theorem \citep{kakutani1948equivalence}. The two measures are equivalent if the following sum converges
\begin{equation*}
    \sum_{j=1}^{\infty} \log \int_{R} \sqrt{\frac{\gamma_{j}(h_j, b_j)}{\gamma_j(0, b_j)}  } \gamma_j(0, b_j)(dx).
\end{equation*}
This leads to the applicable space for $h$ as 
\begin{equation*}
    \mcH_{C}^{*} = \left\{  h\in \mathbb{H} : \sum_{j=1}^{\infty} \left[ \frac{|h_j|}{2b_j} - log(1+ \frac{|h_j|}{2b_j})  \right] < \infty  \right\}.
\end{equation*}
To prove $\mcH_{C}^{*} = \mcH_{C}$, we only need to show that for a non-negative sequence $\{a_j\}_{j\geq 1}$, the series $\sum_{j= 1}^{\infty} [ a_n  - log(1+a_n)]$ converges if and only if $\sum_{j= 1}^{2} a_n^2$ converges. Let $f(x) = x - log(1+x)$ and $g(x) = x^2$. Besides, note that $\lim_{n\rightarrow \infty}f(a_n) = 0$ if and only if $\lim_{n\rightarrow \infty}a_n = 0$. Thus,
\begin{itemize}
    \item If $\sum_{j=1}^{\infty} [ a_n  - log(1+a_n)]<\infty$, then $\lim_{n\rightarrow \infty}f(a_n) = 0$ and so $\lim_{n\rightarrow \infty}a_n = 0$. Then,
    \begin{equation*}
        \lim_{n\rightarrow \infty} \frac{f(a_n)}{g(a_n)} = \lim_{n\rightarrow \infty} \frac{a_n - log(1+a_n)}{a_n^2} = \frac{1}{2}
    \end{equation*}
    therefore by the limit comparison test $\sum_{j=1}^{\infty} a_n^2$ converges too.
    
    \item If $\sum_{j=1}^{\infty} a_n^2<\infty$, by the same statement as above also holds and therefore $\sum_{j=1}^{\infty} [ a_n  - log(1+a_n)]<\infty$.
\end{itemize}
For the ``only if" part, the proof is the same as Theorem 2 in \citet{reimherr2019elliptical}.

\end{proof}

\subsection{Proof of Theorem \ref{thm: density}}\label{thm: density appendix}

\begin{proof}
We prove Theorem~\ref{thm: density} by showing the form of Radon-Nikodym derivative of $\{P_{D}:D\in\mathcal{D}\}$ with respect to $P_{0}$ is the same as derivative of $\gamma(h, C)$ with respect to $\gamma(0, C)$, and it takes the form of 
\begin{equation}\label{eqn: RN derivative in proof}
    \frac{d P_h }{d P_{0} }(z) = \exp\left\{ -\frac{1}{\sigma} \left( \left\|z - h\right\|_{1,C} - \left\| z \right\|_{1,C} \right) \right\}.
\end{equation}
First, we need to show that the right-hand side of Equation (\ref{eqn: RN derivative in proof}) is well-defined when $h\in \mcH_{1,C}$. Define 
\begin{equation*}
    H_{M}(z) = \sum_{j=1}^{M}\frac{|z_j - h_j | - |z_j|}{b_j} \quad \textit{and} \quad H(z) = \lim_{M\rightarrow \infty} H_{M}(z).
\end{equation*}
We show there exists a set $A$ with $P_{0}(A) = 1$ such that $H(z)$ exists and is finite on $A$. Suppose $z_j \sim \Lap(0,b_j)$, then
\begin{equation*}
\begin{aligned}
    \Var\left(H_{M}(z) \right) & = \sum_{j=1}^{M} \Var \left( \frac{|z_j - h_j | - |z_j|}{b_j} \right) \\
    & = \sum_{j=1}^{M} 3b_{j}^{2} - b_{j}^{2}\exp\left\{ -\frac{2|h_{j}|}{b_{j}}\right\}  -\exp\left\{-\frac{|h_{j}|}{b_{j}}\right\} (b_{j}^{2} - 4b_{j}|h_{j}|) \\
    & = \sum_{j=1}^{M} -b_{j}^2 \left( 1 + \exp\left\{- \frac{|h_{j}|}{b_{j}} \right\} \right)^2 + 4b_{j}^{2}\left( 1 + \frac{|h_{j}|}{b_{j}}\exp\left\{ - \frac{|h_{j}|}{b_{j}} \right\} \right) \\
    & = \sum_{j=1}^{M} b_{j}^{2} \left( 4  - \left( 1 + \exp\left\{- \frac{|h_{j}|}{b_{j}} \right\} \right)^2 \right) + 4b_{j}^{2}  \frac{|h_{j}|}{b_{j}}\exp\left\{ - \frac{|h_{j}|}{b_{j}} \right\}.
\end{aligned}
\end{equation*}
The second equality is based on the following facts that $E |z_{j} - h_{j}| = |h_{j}| + \exp\{ -|h_{j}|/b_{j} \}$, $\E|z_{j}| = b_{j}$ and $\Cov(|z_{j} - h_{j}|,|z_{j}|) = |h_{j}|b_{j} + \exp\left\{ -|h_{j}|/b_{j} \right\} ( b_{j}^2 + b_{j}|h_{j}|/2 )$.
By Fatou's Lemma and the condition $h\in \mcH_{1,C}$, we have $\Var(H(z))< \infty$. The set $A$ is $\Omega$ and with $P_0$-measure 1. Therefore, if $h \in \mcH_{1,C}$, the right-hand side of Equation (\ref{eqn: RN derivative in proof}) exists and is well-defined.

Second, we show that (\ref{eqn: RN derivative in proof}) is the Radon-Nikodym derivative of $h + \sigma Z$ with respect to $\sigma Z$. Define
\begin{equation*}
    g(x) = \exp\left\{ -\frac{\sqrt{2}}{\sigma} \left( \left\|x - h\right\|_{1,C} - \left\| x \right\|_{1,C} \right) \right\} \quad \text{and} \quad d P_h^{*} (x) = g(x) d P_0(x).
\end{equation*}
Then, we only need to show that $P_h$ and $P_h^{*}$ are the same probability measure. We accomplish this by showing they have the same moment-generating function.
\begin{equation*}
\begin{aligned}
    MGF_{P_h}(t) & = \E_{P_h} \exp\left\{ \langle X, t \rangle_{\mathbb{H}} \right\} \\
    & = \prod_{j=1}^{\infty} \int_{\mathbb{R}} \exp\left\{ x_j t_j \right\} d \gamma_{h_j,b_j}(x_j) \\
    & = \prod_{j=1}^{\infty} \frac{ \exp\left\{ h_j t_j \right\} }{ 1 - (b_{j} t_j)^{2} }.
\end{aligned}
\end{equation*}
where the second inequality comes from the result that $P_h$ is product measure of $\gamma_{h_j,b_j}$. For the moment generate function of $P_h^{*}$,
\begin{equation*}
\begin{aligned}
    MGF_{P_h^{*}}(t) & = \E_{P_h^{*}} \exp\left\{ \langle X, t \rangle_{\mathbb{H}} \right\} \\
    & = \E_{Q} g(X) \exp\left\{ \langle X, t \rangle_{\mathbb{H}} \right\} \\
    & = \prod_{j=1}^{\infty} \int_{\mathbb{R}} \exp \left\{ -\frac{1}{\sigma}\left( \frac{|x_j - h_j| - |x_j| }{b_j} \right) + x_j t_j \right\}d \gamma_{0,b_j}(x_j) \\
    & = \prod_{j=1}^{\infty} \int_{\mathbb{R}} \exp\left\{ x_j t_j \right\} d \gamma_{h_j,b_j}(x_j) \\
    & = \prod_{j=1}^{\infty} \frac{ \exp\left\{ h_j t_j \right\} }{ 1 - (b_{j} t_j)^{2} } = MGF_{P_h}(h).
\end{aligned}
\end{equation*}
Therefore, $P_h$ and $P_h^{*}$ have the same moment-generating functions and thus are the same probability measure.
\end{proof}

\subsection{Proof of Theorem \ref{thm: gap theorem}}
\begin{proof} 
We prove the theorem via contradiction. Assume if $f_{D} \in \mcH_{C}\backslash \mcH_{1,C}$, for any given fixed $\epsilon$, there exists a $\sigma \in \mathbb{R}^{+}$ such that the ICLP mechanism release $f_D + \sigma Z$ still satisfies $\epsilon$-DP. Then, by the post-processing property of differential privacy, for any transformation $G: \mcH \rightarrow \mcH$, $G(f_D)$ is also $\epsilon$-DP. Now, for any $J\in \mathbb{N}$, consider $G$ to be a projection mapping into first $J$ components, i.e., $G_{J}(f_D) = \sum_{j=1}^{J}\langle f_D,\phi_j \rangle \phi_j$. Therefore, for any $J\in \mathbb{N}$ and $f_{D} \in \mcH_{C}\backslash \mcH_{1,C}$, $G_{J}(f_D)$ is $\epsilon$-DP, i.e.,
\begin{equation*}
    \exp\left\{ - \frac{\sqrt{2}}{\sigma} \sum_{j=1}^{J} \left( \frac{| \langle f_{D} - z,\phi_j \rangle|}{\sqrt{2} b_j} - \frac{| \langle  z,\phi_j \rangle|}{\sqrt{2} b_j} \right) \right\} \leq \exp\left\{ \epsilon \right\},
\end{equation*}
except for $z\in A$ where $A$ is zero-measure set.

Define $B_{j} = \left\{ z_j : \left| z_j \right| > \left| \langle f_{D},\phi_j \rangle \right| \right\}$ and $S_{J} = \left\{ z \in \ell^{2} : z_{j} \in B_{j}, \forall 1\leq j \leq J \ \textit{and} \ z_{j} \in \mathbb{R},  \forall  j > J \right\}$. Then for all $z\in S_{J}$,
\begin{equation*}
     \exp\left\{ - \frac{\sqrt{2}}{\sigma} \sum_{j=1}^{J} \left( \frac{| \langle f_{D} - z,\phi_j \rangle|}{\sqrt{2} b_j} - \frac{| \langle  z,\phi_j \rangle|}{\sqrt{2} b_j} \right) \right\} = \exp\left\{  \frac{\sqrt{2}}{\sigma} \sum_{j=1}^{J} \left( \frac{| \langle f_{D} ,\phi_j \rangle|}{\sqrt{2} b_j}  \right) \right\} \leq \exp\left\{ \epsilon \right\}.
\end{equation*}
However, since $h \in \mcH_{C} \backslash \mcH_{1,C}$, for any given fixed $\epsilon$, one can always find an $J$ so that 
\begin{equation*}
    \exp\left\{ \frac{1}{\sigma} \sum_{j=1}^{J}\frac{|\langle f_{D} ,\phi_j \rangle|}{b_j} \right\} > \exp\left\{ \epsilon \right\}
\end{equation*}
and therefore contradiction holds and no such $\sigma \in \mathbb{R}^{+}$ exists.

The remaining thing is to prove that $S_{J}$ is not a zero-measure set. By the Existence of Product Measure in \citet{tao2011introduction}, 
\begin{equation*}
    \gamma_{0,C}(S_J) = \prod_{j=1}^{J}\gamma_{0,\sqrt{2}b_j}(B_j)
\end{equation*}
where the right-hand side of the equation is greater than $0$ by definition of $B_j$.
\end{proof}

\subsection{Proof of Theorem \ref{thm: the ICLP mechanism theorem}}
\begin{proof}
By Theorem~\ref{thm: density}, the density of $\tilde{f}_{D}$ with respect to to $\sigma Z$ is
\begin{equation*}
    \frac{d P_D }{d P_{0} }(z) = \exp\left\{ -\frac{1}{\sigma} \left( \left\|z - f_{D}\right\|_{1,C} - \left\| z \right\|_{1,C} \right) \right\}.
\end{equation*}
We aim to show that for any measurable subset $A\subseteq \mbH$, one has
\begin{equation*}
    P_{D}(A) \leq e^{\epsilon} P_{D'}(A),
\end{equation*}
which is equivalent to show
\begin{equation*}
    P_{D}(A) = \int_{A} d P_{D}(x) = \int_{A} \frac{d P_{D}}{d P_{D'}}(x) d P_{D'}(x) \leq e^{\epsilon} \int_{A} d P_{D'}(x).
\end{equation*}
Notice
\begin{equation*}
\begin{aligned}
    \frac{d P_{D}}{d P_{D'}}(x) & = \frac{d P_{D}}{d P_0}(x) /\frac{d P_{D'}}{d P_0}(x) \\
    & = \exp\left\{ -\frac{1}{\sigma} \left( \left\|x - f_{D}\right\|_{1,C} - \left\| x - f_{D'}\right\|_{1,C} \right) \right\} \\
    & \leq \exp\left\{ \frac{1}{\sigma}  \left\|f_{D'} - f_{D}\right\|_{1,C}   \right\}.
\end{aligned}
\end{equation*}
Recall the global sensitivity for the ICLP mechanism is defined as 
\begin{equation*}
    \Delta = \sup_{D\sim D'} \|f_{D} - f_{D'}\|_{1,C}.
\end{equation*}
By taking $\sigma = \frac{\Delta}{\epsilon}$, we have $\frac{d P_{D}}{d P_{D'}}(x)\leq e^{\epsilon}$, $\forall x \in \mbH$. Thus, the desired inequality holds, i.e.,
\begin{equation*}
    P_{D}(A)  = \int_{A} \frac{d P_{D}}{d P_{D'}}(x) d P_{D'}(x) \leq e^{\epsilon} \int_{A} d P_{D'}(x).
\end{equation*}
\end{proof}

\subsection{Proof of Theorem \ref{thm: continuity ICLP theorem}}
\begin{proof}
First, we decompose the ICLP as $Z(t) - Z(s) = \sum \lambda_j^{1/2} Z_j (\phi_j(t) - \phi_j(s))$, which leads to
\begin{equation*}
\E[\exp\{ t (Z(t) - Z(s))  \}]
 =\prod_{j=1}^{\infty} \frac{1}{1-t^2\lambda_j(\phi_j(t) - \phi_j(s))^2}
= \exp\left\{
-\sum_{j=1}^{\infty} \log\left(1 - \frac{t^2\lambda_j}{2}(\phi_j(t) - \phi_j(s))^2 \right)
\right\}
\end{equation*}
with $t$ satisfies $0 \leq t^2\lambda_j(\phi_j(t) - \phi_j(s))^2 < 1$ for all $j$. Let $C_{t} = C(t,\cdot)$ for any $t\in T$, then by the $\alpha$-H\"{o}lder continuous property of $C$, we have
\begin{equation*}
\begin{aligned}
\lambda_j(\phi_j(t) - \phi_j(s))^2 
& = \lambda_j \langle C_t - C_s, \phi_j \rangle_{C}^2 \\
& \leq \langle C_t - C_s,   C_t - C_s \rangle_{C}  \\
& = C(t,t) -2C(t,s) + C(s,s) \\
& \leq  2M_C | t- s|^{\alpha},
\end{aligned}
\end{equation*}
where $M_C$ is the H\"older-continuous constant, and this leads to
\begin{equation*} 
    0 \leq t \leq (\frac{1}{M_C })^{\frac{1}{2}} |t-s|^{-\frac{\alpha}{2}}.
\end{equation*}
For $x\in [0, 1)$, define function $f(x) = -\log(1-x)$, by the mean value theorem, we have
\begin{equation*}
    -\log(1-x) = f(x) = f(0) + x f'(\zeta) = \frac{x}{1-\zeta}  \leq \frac{x}{1-x},
\end{equation*}
for some $\zeta \in (0, x)$. Therefore, applying the above inequality, we have
\begin{equation*}
\begin{aligned}
-\log(1 - \frac{t^2\lambda_j}{2}(\phi_j(t) - \phi_j(s))^2)
& \leq \frac{\frac{t^2\lambda_j}{2}(\phi_j(t) - \phi_j(s))^2}{1 - \frac{t^2\lambda_j}{2}(\phi_j(t) - \phi_j(s))^2}\\
& \leq \frac{t^2\lambda_j}{2}(\phi_j(t) - \phi_j(s))^2
\max_{k} \left\{ (1 - \frac{t^2\lambda_k}{2}(\phi_k(t) - \phi_k(s))^2)^{-1} \right\}.
\end{aligned}
\end{equation*}
Choosing $t$ such that $(1 - \frac{t^2\lambda_k}{2}(\phi_k(t) - \phi_k(s))^2 )^{-1}\leq M_0< \infty$, we have
\begin{equation*}
\begin{aligned}
\E[\exp\{ t (Z(t) - Z(s))  \}]
& \leq \exp\left\{ M_0 \frac{t^2}{2} \sum\lambda_j(\phi_j(t) - \phi_j(s))^2  \right\} \\
& = \exp\left\{  \frac{M_0 t^2}{2} (C(t,t)- 2C(t,s) + C(s,s) )\right\} \\
& \leq \exp\left\{ M_0 M_C t^2 |t-s|^{\alpha} \right\}.
\end{aligned}
\end{equation*}
By the Chernoff bound, we have
\begin{equation*}
\begin{aligned}
    P\left( |Z(t) - Z(s)| \geq a \right) &\leq \frac{\E[\exp\{ t (Z(t) - Z(s))  \}]}{\exp\{ t a \}} \\
    & \leq  \exp\left\{M_0 M_C t^2  |t-s|^{\alpha} - t a\right\} .
\end{aligned}
\end{equation*}
The minimizer of the right-hand with respect to $t$ is $t_0 = \frac{a}{2M_0 M_C |t-s|^{\alpha}}$. With restriction of $t$, we get $a \leq 2 M_0 |t-s|^{\frac{1}{2}\alpha}$.

We consider the following two cases:

\textbf{Case 1 :} Suppose $a \leq 2 M_0 |t-s|^{\frac{1}{2}\alpha}$, the minimizer is $t_0 = \frac{a}{2M_0 M_C |t-s|^{\alpha}}$. We have 
\begin{equation*}
    P\left( |Z(t) - Z(s)| \geq a \right) 
    \leq  \exp\left\{ - \tilde{M_1} a^2 |t-s|^{-\alpha} \right\},
\end{equation*}
for some generic constant $\tilde{M}_1$. Define $a(x) = C |x|^{\beta}$ with $\beta \geq   \frac{1}{2}\alpha$, and two series as 
\begin{equation*}
    \sum_{n=1}^{\infty} a(2^{-n}) = \sum_{n=1}^{\infty}  2^{-n\beta}  \quad \textit{and } \quad \sum_{n=1}^{\infty} 2^{n} \exp\{- \tilde{M}_2 |2^n|^{\alpha-2\beta} \}.
\end{equation*}
The first series converges if $\beta <1$. However, to make the second one converge, we need $\alpha > 2\beta$, which leads to $\alpha > \alpha$ contradiction. 

\textbf{Case 2 :} Suppose $a > 2 M_0 |t-s|^{\frac{1}{2}\alpha}$, then the minimizer is $t_0 = (\frac{1}{M_C })^{\frac{1}{2}} |t-s|^{-\frac{1}{2}\alpha}$, then 
\begin{equation*}
    \exp\left\{M_0 M_C t^2  |t-s|^{\alpha} - t a\right\}  = \exp\left\{ M_0 - \left( \frac{1}{M_C } \right)^{\frac{1}{2}} |t-s|^{-\frac{1}{2}\alpha} a  \right\}.
\end{equation*}
By picking function $a(x) = 2M_0 |x|^{\beta} > 2M_0|x|^{\frac{1}{2}\alpha}$, with $\beta\in (0,\frac{1}{2}\alpha)$, we have $\sum_{j=1}^{\infty} a(2^{-n}) < \infty$ and 
\begin{equation*}
\begin{aligned}
    \sum_{j=1}^{\infty} 2^{n} b(2^{-n}) & := \sum_{j=1}^{\infty} 2^{n} \exp\left\{ M_0 - \left( \frac{1}{M_C } \right)^{\frac{1}{2}} |2^{-n}|^{-\frac{1}{2}\alpha}  a(2^{-n})  \right\} \\
    & = \sum_{j=1}^{\infty} 2^{n} \exp\left\{ M_0 - \left( \frac{1}{M_C } \right)^{\frac{1}{2}} |2^n|^{\frac{1}{2}\alpha -\beta}  \right\}. \\
\end{aligned}
\end{equation*}
Since $\beta\in (0,\frac{1}{2}\alpha)$, we have $\sum_{j=1}^{\infty} 2^{n} b(2^{-n}) < \infty$. 

Therefore, combining the results from both cases, we have 
\begin{equation*}
    \sum_{n=1}^{\infty} 2^{n} P\left( |Z(t+2^{-n}) - Z(t)| \geq 2^{-n} \right) < \infty.
\end{equation*}
Finally, the rest of the proof follows the same proof of Theorem 5.2.8 in \citet{lunardi2015infinite}.
\end{proof}

\section{Proofs for Section \ref{sec: methodology section}}

\subsection{Proof of Theorem~\ref{thm: GS analysis for mean protection}}
\begin{proof}\label{proof: GS analysis for mean protection}
For the FRL, ICLP-AR, and ICLP-QR mechanisms, we first derive their exact solution and then conduct the global sensitivity analysis. For two $n$ sample adjacent datasets $D$ and $D'$, we assume they only differ in the first observation, i.e., $X_{1}$ and $X_{1}'$.

\paragraph{FRL:}
For FRL, the estimator can be expressed as 
\begin{equation*}
    \hat{f}_{D} = \sum_{j=1}^{M} \langle \bar{X}, \phi_{j} \rangle_{\mbH} \phi_{j}
\end{equation*}
Its private version can be obtained by applying the multivariable version of the Laplace mechanism to the coefficients $\hat{f}_{D,1:M} = (\langle \bar{X}, \phi_{1} \rangle_{\mbH},\cdots, \langle \bar{X}, \phi_{M} \rangle_{\mbH})$, i.e., 
\begin{equation*}
    \hat{f}_{D} = \sum_{j=1}^{M} \left\{ \langle \bar{X}, \phi_{j} \rangle_{\mbH}  + Z_{j} \right\} \phi_{j}, \quad \text{with} \quad Z_{j} \stackrel{i.i.d.}{\sim} Lap\left(0, \frac{\Delta}{\epsilon}\right)
\end{equation*}
where $\Delta = \sup_{D\sim D'}\|\hat{f}_{D,1:M} - \hat{f}_{D', 1:M}\|_{\ell^{1}}$. Given the bounded norm in Assumption~\ref{assmption: bounding norm},
\begin{equation*}
    \Delta = \sup_{D\sim D'}\|\hat{f}_{D,1:M} - \hat{f}_{D', 1:M}\|_{\ell^{1}} \leq \frac{1}{n}\sum_{j=1}^{M} | \langle X_{1} - X_{1}', \phi_{j} \rangle_{\mbH} | \leq \frac{2M\tau}{n}.
\end{equation*}

\paragraph{ICLP-AR:}
To obtain the closed form of the ICLP-AR estimator, we expand $X_{i} - \theta$ by the eigenfunctions $\phi_{j}$, i.e.
\begin{equation}
\begin{aligned}
    \frac{1}{n}\sum_{i=1}^{n}\left\| X_i - \theta \right\|_{\mbH}^{2} + \psi \|\theta\|_{1,C^{\eta_{l}}} &  =  \frac{1}{n}\sum_{i=1}^{n}\left\| \sum_{j=1}^{\infty} \langle X_i - \theta, \phi_j \rangle_{\mbH} \phi_{j} \right\|_{\mbH}^{2} + \psi \|\theta\|_{1,C^{\eta_{l}}}\\
    & = \frac{1}{n}\sum_{i=1}^{n} \sum_{j=1}^{\infty} \langle X_i - \theta, \phi_j \rangle_{\mbH}^{2} + \psi \|\theta\|_{1,C^{\eta_{l}}} \\
    & = \sum_{j=1}^{\infty} \left\{  \frac{1}{n}\sum_{i=1}^{n}  ( X_{ij} - \theta_{j} )^{2} + \psi \frac{|\theta_{j}|}{\lambda_{j}^{\eta_{l}/2}} \right\}
\end{aligned}
\end{equation}
Solving the minimization problem within the bracket, then for each $j$, we have
\begin{equation*}
    \hat{\theta}_j = \text{sgn} (\bar{X}_j ) \left( \bar{X}_j - \frac{\psi}{2\lambda_j^{\eta_{l}/2}} \right)^{+}.
\end{equation*}
This leads to the ICLP-AR estimator as 
\begin{equation*}
    \hat{\mu}_{D}^{l} = \sum_{j=1}^{\infty} \hat{\theta}_{j} \phi_{j} = \sum_{j=1}^{\infty}s_{\psi, 2\lambda_j^{\eta_{l}/2}}\left( \left\langle \bar{X},\phi_j \right\rangle_{\mbH} \right) \phi_{j}.
\end{equation*}
For the global sensitivity, 
\begin{equation*}
\begin{aligned}
    \sup_{D \sim D'} \| \hat{\mu}_{D}^{l} - \hat{\mu}_{D'}^{l} \|_{1,C} & = \sup_{D \sim D'} \sum_{j = 1}^{J_{\tau}} \frac{ |s_{\psi, \lambda_j^{\eta_{l}/2}}\left( \left\langle \bar{X}_{D},\phi_j \right\rangle_{\mbH} \right) - s_{\psi, \lambda_j^{\eta_{l}/2}}\left( \left\langle \bar{X}_{D'},\phi_j \right\rangle_{\mbH} \right)| }{\lambda_j^{\frac{1}{2}}} \\
    & \leq \sup_{D\sim D'} \sum_{j = 1}^{J_{\tau}} \frac{|\left\langle \bar{X}_{D} - \bar{X}_{D'},\phi_j \right\rangle_{\mbH} |}{\lambda_j^{\frac{1}{2}}} \leq \frac{2\tau}{n} \sum_{j = 1}^{J_{\tau}} \frac{1}{\lambda_j^{\frac{1}{2}}},
\end{aligned}
\end{equation*}
where the first inequality is based on the fact that, in the worst case, the $j$-th coefficients based on $D$ and $D'$ will not be shrunk to $0$ simultaneously and thus should have the same sensitivity without soft-threshold function.

\paragraph{ICLP-QR:}
Recall the object function 
\begin{equation*}
    F(\theta) = \frac{1}{n}\sum_{i=1}^{n}\left\| X_i - \theta \right\|_{\mbH}^{2} + \psi \|\theta\|_{C^{\eta_{r}}}^{2}.
\end{equation*}
and after dropping everything not involving $\theta$, we have 
\begin{equation*}
\begin{aligned}
    F(\theta) & = -2\langle \bar{X}, \theta \rangle_{\mbH} + \langle \theta, \theta \rangle_{\mbH} + \psi \langle \theta, \theta \rangle_{C^{\eta_{r}}} \\
    & = -2\langle \bar{X}, C^{\eta_{r}} \theta \rangle_{C^{\eta_{r}}} + \langle \theta, C^{\eta_{r}}\theta \rangle_{C^{\eta_{r}}} + \psi \langle \theta, \theta \rangle_{C^{\eta_{r}}}.
\end{aligned}
\end{equation*}
The second equality is based on Hilbert space's own dual, i.e., $\langle \cdot, \cdot \rangle_{\mbH} = \langle \cdot, C(\cdot) \rangle_{\mcH_{C}}$. Thus the minimizer of the $F(\theta)$ is 
\begin{equation*}
\begin{aligned}
    \hat{\mu}_{D}^{r} & = \left( C^{\eta_{r}} + \psi I \right)^{-1} C^{\eta_{r}}(\bar{X}) \\
    & = \sum_{j=1}^{\infty} \frac{\lambda_j^{\eta_{r}}}{\lambda_j^{\eta_{r}} + \psi} \left\langle \bar{X},\phi_j \right\rangle_{\mbH} \phi_j
\end{aligned}
\end{equation*}
where the second equality follow by expansion $\hat{\mu}_{D}^{r}$ under the eigenfunction $\phi_{j}$. For the global sensitivity, the upper bound for $\sup_{D\sim D'} \| \hat{\mu}_{D}^{r} - \hat{\mu}_{D'}^{r} \|_{1,C}$ is
\begin{equation*}
    \begin{aligned}
        \sup_{D\sim D'} \| \hat{\mu}_{D} - \hat{\mu}_{D'} \|_{1,C} & = \sup_{D\sim D'} \sum_{j=1}^{\infty} \frac{ \lambda_{j}^{\eta_{r} - \frac{1}{2}} }{  \lambda_{j}^{\eta_{r}} + \psi }  \left|\left\langle \bar{X} - \bar{X}',\phi_j \right\rangle_{\mbH} \right| \\
        & \leq \frac{2\tau}{n} \sum_{j=1}^{\infty} \frac{ \lambda_{j}^{\eta_{r} - \frac{1}{2}} }{  \lambda_{j}^{\eta_{r}} + \psi },
    \end{aligned}
\end{equation*}
where the inequality is based on Cauchy-Schwarz inequality and the $\|X_{i}\|_{\mbH}\leq \tau$. 
\end{proof}

\subsection{Proof of Theorem~\ref{thm: total error for mean protection}}
\begin{proof}

\paragraph{FRL:}
For the privacy error,
\begin{equation*}
    \E \left\| \tilde{\mu} - \hat{\mu} \right\|_{\mbH}^{2} = \frac{2\Delta^2}{\epsilon^2} \cdot M \leq \frac{8\tau^2}{\epsilon^2 n^2} M^3.
\end{equation*}
For the estimation error,
\begin{equation*}
\begin{aligned}
    \E \left\| \hat{\mu} - \mu_{0}\right\|_{\mbH}^{2} & = \frac{1}{n} \sum_{j=1}^{M} \E \langle X_{1} - \mu_{0}, \phi_{j} \rangle_{\mbH}^2 + \sum_{j=M+1}^{\infty} \langle \mu_{0}, \phi_{j} \rangle_{\mbH}^2 \\
    & \lesssim \frac{1}{n} + \sum_{j=M+1}^{\infty} \langle \mu_{0}, \phi_{j} \rangle_{\mbH}^2 \\
    & \lesssim \frac{1}{n} + \lambda_{M}^{\eta}\|\mu_{0}\|_{C^{\eta}}^2 \\
    & \lesssim \frac{1}{n} + M^{-\eta\beta}.
\end{aligned}
\end{equation*}
The first inequality is based on Assumption~\ref{assmption: bounding norm}, while the second one is based on the fact that $\mu_{0}\in \mcH_{C^{\eta}}$.

\paragraph{ICLP-AR:}
For the privacy error,
\begin{equation*}
\begin{aligned}
    \E \left\| \tilde{\mu} - \hat{\mu} \right\|_{\mbH}^{2} & = \frac{\Delta^2}{\epsilon^2} \sum_{j=1}^{J^{*}}\lambda_{j} \\
    & \lesssim \frac{4\tau^2}{\epsilon^2 n^2} \left( \sum_{j=1}^{J^{*}} j^{\frac{\beta}{2}} \right)^1 \left( \sum_{j=1}^{\infty} j^{-\beta} \right) \\
    & \lesssim \frac{1}{\epsilon^2 n^2} (J^{*})^{\beta + 2}
\end{aligned}    
\end{equation*}
Next, we turn to estimation error. Define $\mu_{0,\psi} = \sum_{j=1}^{\infty} f_{\psi,2\lambda_{j}^{\frac{\eta_{l}}{2}}}\left( \left\langle \mu_0, \phi_j \right\rangle \right)\phi_j$, by triangular inequality
\begin{equation*}
    \E \left\| \hat{\mu} - \mu_{0} \right\|_{\mbH}^{2} \leq \E \left\| \hat{\mu} - \mu_{0,\psi} \right\|_{\mbH}^{2} + \left\| \mu_{0,\psi} - \mu_{0} \right\|_{\mbH}^{2}.
\end{equation*}
For the bias term, let $A = \left\{j : |\mu_j|\geq \frac{\psi}{2\lambda_{j}^{\eta_{l}/2}} \right\}$, then
\begin{equation*}
    \left\| \mu_{0,\psi} - \mu_{0} \right\|_{\mbH}^{2} = \sum_{A} \left( \mu_{0j} - f_{\psi,2\lambda_{j}^{\frac{\eta_{l}}{2}}}( \mu_{0j} ) \right)^2 + \sum_{A^c} \left( \mu_{0j} - f_{\psi,2\lambda_{j}^{\frac{\eta_{l}}{2}}}( \mu_{0j} ) \right)^2.
\end{equation*}
Starting with the summation over $A$, since $\frac{\lambda_{j}^{-\frac{\eta_{l}}{2}}}{2} < \frac{|\mu_{0j}|}{\psi}$ we have
\begin{equation*}
\begin{aligned}
    \sum_{A} \left( \mu_{0j} - f_{\psi,2\lambda_{j}^{\frac{\eta_{l}}{2}}}( \mu_{0j} ) \right)^2 & = \sum_{A} \frac{\psi^2}{4\lambda_{j}^{\eta_{l}}}  \leq \psi \sum_{A}\frac{\left| \mu_{0j} \right|}{2\lambda_{j}^{\frac{\eta_{l}}{2}}} \leq \frac{\psi }{2}\left\| \mu_{0} \right\|_{1,C^{\eta_{l}}}.
\end{aligned}
\end{equation*}
Turning to summation over $A^c$, since $|\mu_{0j}| \leq \frac{\psi}{2\lambda_{j}^{\frac{\eta_{l}}{2}}}$,
\begin{equation*}
\begin{aligned}
     \sum_{A^c} \left( \mu_{0j} - f_{\psi,2\lambda_{j}^{\frac{\eta_{l}}{2}}}( \mu_{0j} ) \right)^2 & = \sum_{A^c} \mu_{0j}^2  \leq \psi \sum_{A^c} \frac{|\mu_{0j}|}{2\lambda_{j}^{\frac{\eta_{l}}{2}}} \leq \frac{\psi}{2} \left\| \mu_{0} \right\|_{1,C^{\eta_{l}}}.
\end{aligned}
\end{equation*}
Therefore, the overall bias is bounded by 
\begin{equation*}
    \left\| \mu_{0,\psi} - \mu_{0} \right\|_{\mbH}^{2} \leq \frac{\psi}{2} \left\| \mu_{0} \right\|_{1,C^{\eta_{l}}}.
\end{equation*}
Now consider variance term $\E \left\| \hat{\mu} - \mu_{0,\psi} \right\|_{\mbH}^{2}$, 
\begin{equation*}
    \left\| \hat{\mu}  - \mu_{0} \right\|_{\mbH}^{2} = \sum_{j=1}^{\infty} \left( f_{\psi,2\lambda_{j}^{\frac{\eta_{l}}{2}}}\left( \bar{X}_j \right) - f_{\psi,2\lambda_{j}^{\frac{\eta_{l}}{2}}}\left( \mu_{0j} \right) \right)^2.
\end{equation*}
Similar to the bias part, the summation can be decomposed into the sum of four disjoint pieces
\begin{align*}
    A_{0,0} &= \{ | \bar X_j| \leq \psi/2\lambda_{j}^{\frac{\eta_{l}}{2}}, | \mu_{j}| \leq \psi/2\lambda_{j}^{\frac{\eta_{l}}{2}} \}, \\
    A_{0,1} &= \{ | \bar X_j| \leq \psi/2\lambda_{j}^{\frac{\eta_{l}}{2}}, | \mu_{j}| > \psi/2\lambda_{j}^{\frac{\eta_{l}}{2}} \}, \\
    A_{1,0} &= \{ | \bar X_j| > \psi/2\lambda_{j}^{\frac{\eta_{l}}{2}}, | \mu_{j}| \leq \psi/2\lambda_{j}^{\frac{\eta_{l}}{2}} \}, \\
    A_{1,1} &= \{ | \bar X_j| > \psi/2\lambda_{j}^{\frac{\eta_{l}}{2}}, | \mu_{j}| > \psi/2\lambda_{j}^{\frac{\eta_{l}}{2}} \}. 
\end{align*}
When $j\in A_{0,0}$, the summation is zero. Consider $j\in A_{0,1}$, since $|\bar{X}_{j}| \leq \frac{\psi}{2\lambda_{j}^{\frac{\eta_{l}}{2}}}$, 
\begin{equation*}
\begin{aligned}
    \left( f_{\psi,2\lambda_{j}^{\frac{\eta_{l}}{2}}}\left( \bar{X}_j \right) - f_{\psi,2\lambda_{j}^{\frac{\eta_{l}}{2}}}\left( \mu_{0j} \right) \right)^2  = \left( f_{\psi,2\lambda_{j}^{\frac{\eta_{l}}{2}}}\left( \mu_{0j} \right)  \right)^2 = \left( \mu_{0j} - \operatorname{sgn}(\mu_{0j})\frac{\psi}{2\lambda_{j}^{\frac{\eta_{l}}{2}}}  \right)^2  \leq \left( \mu_{0j} - \bar{X}_{j} \right)^2.
\end{aligned}
\end{equation*}
By symmetry, we get the same bound over $A_{1,0}$.  So lastly we consider summation over $A_{1,1}$  For $j \in A_{1,1}$ we have
\begin{equation*}
\begin{aligned}
        \left( f_{\psi,2\lambda_{j}^{\frac{\eta_{l}}{2}}}\left( \bar{X}_j \right) - f_{\psi,2\lambda_{j}^{\frac{\eta_{l}}{2}}}\left( \mu_{0j} \right) \right)^2 & = \left( \mu_{0j} - \bar{X}_j - \left( \operatorname{sgn}(\mu_{0j}) - \operatorname{sgn}(\bar{X}_j)  \right)\frac{\psi}{2\lambda_{j}^{\frac{\eta_{l}}{2}}} \right)^2.
\end{aligned}
\end{equation*}
If both $\mu_{0j}$ and $\bar X_j$ have the same sign, then this is just $(\mu_{0j} - \bar X_j)^2$.  If they have opposite signs, then we have
\begin{equation*}
     \left| \left(\operatorname{sgn}(\mu_{0j}) - \operatorname{sgn}(\bar{X}_j) \right) \frac{\psi}{2\lambda_{j}^{\frac{\eta_{l}}{2}}} \right| \leq \left|\mu_{0j} - \bar{X}_j \right|.
\end{equation*}
Therefore, 
\begin{equation*}
    \left( f_{\psi,2\lambda_{j}^{\frac{\eta_{l}}{2}}}\left( \bar{X}_j \right) - f_{\psi,2\lambda_{j}^{\frac{\eta_{l}}{2}}}\left( \mu_{0j} \right) \right)^2 \leq 4 \left( \mu_{0j} - \bar{X}_{j} \right)^2, \quad \textit{for} \quad j\in A_{1,1}.
\end{equation*}
Finally, the overall variance term is bounded by
\begin{equation*}
    \E \left\| \hat{\mu}  - \mu_{0} \right\|_{\mbH}^{2} \leq 4 \E \left\| \bar{X}  - \mu_{0} \right\|_{\mbH}^{2} \leq \frac{4}{n}\E \left\| X_{1} \right\|_{\mbH}^2 \lesssim n^{-1}.
\end{equation*}

\paragraph{ICLP-QR:}
Recall the global sensitivity,
\begin{equation*}
        \sup_{D\sim D'} \| \hat{\mu}_{D} - \hat{\mu}_{D'} \|_{1,C}\leq \frac{2\tau}{n} \sum_{j=1}^{\infty} \frac{ \lambda_{j}^{\eta_{r} - \frac{1}{2}} }{  \lambda_{j}^{\eta_{r}} + \psi }.
\end{equation*}
For the summation term, observe that given $\eta > \frac{1}{2} + \frac{1}{\beta}$
\begin{equation*}
\begin{aligned}
    \sum_{j=1}^{\infty} \frac{ \lambda_{j}^{\eta_{r} - \frac{1}{2}} }{  \lambda_{j}^{\eta_{r}} + \psi } & \leq \int_{0}^{\infty} \frac{C_{1}x^{\frac{\beta}{2}}}{C_{2} + x^{\eta \beta}} dx \cdot \psi^{-\frac{1}{\eta} \left( \frac{1}{\beta} + \frac{1}{2} \right)} \\
    & \leq \left\{ \int_{0}^{M} \frac{C_{1}x^{\frac{\beta}{2}}}{C_{2} + x^{\eta \beta}} dx + \int_{M}^{\infty} x^{\frac{\beta}{2} - \eta \beta} dx \right\}\cdot \psi^{-\frac{1}{\eta} \left( \frac{1}{\beta} + \frac{1}{2} \right)} \\
    &\lesssim \psi^{-\frac{1}{\eta} \left( \frac{1}{\beta} + \frac{1}{2} \right)},
\end{aligned}
\end{equation*}
where the first inequality is based on Assumption~\ref{assump: EDR} and change of variable. Therefore, for privacy error, we have
\begin{equation*}
    \E \|\tilde{\mu} - \hat{\mu}\|_{\mbH}^{2}  = \frac{\Delta^2}{\epsilon^2}\sum_{j=1}^{\infty}\lambda_j \lesssim (n\epsilon)^{-2} \cdot \psi^{-\frac{2}{\eta}\left( \frac{1}{\beta} + \frac{1}{2}\right)} .
\end{equation*}
For estimation error, the $n^{-1}$ part comes from variance while for bias, we have
\begin{equation*}
    \|\E\hat{\mu} - \mu_0\|_{\mbH}^{2} = \sum_{j=1}^{\infty}\left( \frac{\psi}{\lambda_j^{\eta_{r}}+\psi} \right)^2 \left\langle \mu_0,\phi_j \right\rangle^2 \leq \psi \| \mu_0\|_{C^{\eta_{r}}}^2 \lesssim \psi,
\end{equation*}
where the last inequality is by assuming $\| \mu_0\|_{C^{\eta_{r}}}<\infty$. Combining privacy error and estimation error, one gets the desired results.
\end{proof}

\subsection{Proof of Theorem \ref{thm:kde gs and utility}}
\begin{proof}
Recall that the exact form of the kernel density estimator is
\begin{equation*}
    \hat{K}_{D}(x) = \frac{1}{n \sqrt{det(\mathbf{H})}} \sum_{i=1}^{n} K^{\eta} \left( \mathbf{H}^{-\frac{1}{2}} (x-x_{i}) \right).
\end{equation*}
Then, by the definition of global sensitivity, we have
\begin{equation*}
\begin{aligned}
    \Delta = \sup_{D\sim D'} \left\| \hat{K}_{D} - \hat{K}_{D'} \right\|_{1,K}
    & \leq \frac{1}{n \sqrt{det(\mathbf{H})} } \left\| K^{\eta}(\mathbf{H}^{-\frac{1}{2}} x_n) - K^{\eta}(\mathbf{H}^{-\frac{1}{2}} x_n^{'})  \right\|_{K^{\eta}} \sqrt{tr(K^{\eta-1})} \\
    & \leq \frac{1}{n \sqrt{det(\mathbf{H})} } \sqrt{tr(K^{\eta-1})} \sqrt{ 2 \left( K^{\eta} (0) - K^{\eta} ( \mathbf{H}^{-\frac{1}{2}} (x_n - x_n^{'}) ) \right)  } \\
    & \leq \frac{2M_K}{n \sqrt{det(\mathbf{H})} } \sqrt{tr(K^{\eta-1})}.
\end{aligned}
\end{equation*}
The first inequality is based on the Cauchy–Schwarz inequality, which is also used in deriving the ICLP-QR strategy. The last inequality holds by the assumption that $K^{\eta}(\cdot,\cdot)$ is point-wise bounded. 

Turning to the utility, taking $\mathbf{H}$ to be a diagonal matrix with the same entry, then by the assumptions stated in the Theorem \ref{thm:kde gs and utility} and by the Theorem 6.28 in \citet{wasserman2006all}, the risk $R$ satisfies
\begin{equation*}
\begin{aligned}
     R & = \E \int_{T} \left( \tilde{f}_{D}(x) - f_0(x) \right)^2 dx\\
     & \leq 2* \left( \E \int_{T} \left( \tilde{f}_{D}(x) - \hat{f}_{D}(x) \right)^2 dx + \E \int_{T} \left( \hat{f}_{D}(x) - f_0(x) \right)^2 dx  \right)\\
     & \leq O\left( \frac{c_{1}}{n^2 h^{2d}} + h^4 + \frac{c_{2}}{n h^d} \right).
\end{aligned}
\end{equation*}
for some constants $c_1$ and $c_2$.
\end{proof}

\subsection{Proof of Theorem \ref{thm:GS for ERM}}
\begin{proof}
Recall while deriving the ICLP-QR strategy, for a given $\eta>1$ such that $tr(C^{\eta-1})$ is finite, we have 
\begin{equation*}
    \|h\|_{1,C} \leq \left\| h \right\|_{C^{\eta}} \sqrt{\operatorname{trace}(C^{\eta-1})}.
\end{equation*}
Substituting $h$ by $\hat{f}_{D} - \hat{f}_{D'}$ leads to 
\begin{equation*}
    \|\hat{f}_{D} - \hat{f}_{D'}\|_{1,C} \leq \left\| \hat{f}_{D} - \hat{f}_{D'} \right\|_{C^{\eta}} \sqrt{\operatorname{trace}(C^{\eta-1})},
\end{equation*}
meaning that we need to found the upper bound for $\| \hat{f}_{D} - \hat{f}_{D'} \|_{C^{\eta}}$. Let $t\in [0,1]$, $\delta_{D',D} = \hat{f}_{D'} - \hat{f}_{D}$ and $L_{D}(f) = \frac{1}{n} \sum_{i=1}^{n}L_{d_i, f}$. Note that $\hat{f}_{D'}$ and $\hat{f}_{D}$ are the minimizers of (\ref{eqn:empirical minimium}), we have 

\begin{equation*}
    L_D\left(\hat{f}_{D}\right)+\psi \left\|\hat{f}_{D}\right\|_{C^{\eta}}^2 
     \leq L_D\left(\hat{f}_{D}+ t \delta_{D^{\prime}, D}\right)+\psi \left\|\hat{f}_{D}+ t \delta_{D^{\prime}, D}\right\|_{C^{\eta}}^2,
\end{equation*}
and 
\begin{equation*}
    L_D\left(\hat{f}_{D'}\right)+\psi \left\|\hat{f}_{D'}\right\|_{C^{\eta}}^2 
    \leq L_D\left(\hat{f}_{D'}- t \delta_{D^{\prime}, D}\right)+\psi \left\|\hat{f}_{D'}- t \delta_{D^{\prime}, D}\right\|_{C^{\eta}}^2.
\end{equation*}
Combining the two inequalities above, we have
\begin{equation*}
\begin{aligned}
    L_D\left(\hat{f}_{D}\right) - & L_D\left(\hat{f}_{D}+ t \delta_{D^{\prime}, D}\right)  + L_D\left(\hat{f}_{D'}\right) - L_D\left(\hat{f}_{D'}- t \delta_{D^{\prime}, D}\right)\\ & \leq \psi \left( \left\|\hat{f}_{D}+ t \delta_{D^{\prime}, D}\right\|_{C^{\eta}}^2 -  \left\|\hat{f}_{D}\right\|_{C^{\eta}}^2 + \left\|\hat{f}_{D'}- t \delta_{D^{\prime}, D}\right\|_{C^{\eta}}^2 - \left\|\hat{f}_{D'}\right\|_{C^{\eta}}^2  \right).
\end{aligned}
\end{equation*}
Then, using the same proof techniques in Section 4.3 of \citet{hall2013differential}, we have 
\begin{equation*}
    \|\hat{f}_{D} - \hat{f}_{D'} \|_{C^{\eta}} \leq \frac{M}{\psi n} \sqrt{\sup_{x} C^{\eta}(x,x)},
\end{equation*}
which completes the proof.
\end{proof}

\section{Extension of Mean Function}\label{apd: extension of mean function}
In Section~\ref{sec: mean function protection}, we only considered the mean protection. In this section, we demonstrate that many statistical estimation problems in the context of functional data analysis can be reduced down to mean estimation. Therefore, the mean protection technique and the corresponding theoretical analysis derived in Section~\ref{sec: mean function protection} can be applied to these problems as well. Specifically, we consider the estimation and protection of (1) the covariance function and (2) the coefficient function in function-on-scalar linear regression.

\subsection{Covariance Function}
For a given sample $X_{1},\cdots,X_{n} \in \mbH$, where $\mbH$ represents some function spaces. Our goal is to estimate its covariance function, i.e.,
\begin{equation*}
    C(s,t) = \E \left( X(s) - \mu(s) \right) \left( X(t) - \mu(t) \right)
\end{equation*}
where $\mu$ is the mean function, and the empirical sample covariance function is 
\begin{equation*}
    \bar{C}(s,t) = \frac{1}{n} \sum_{i=1}^{n} \left( X_{i}(s) - \bar{X}(s) \right)\left( X_{i}(t) - \bar{X}(t) \right).
\end{equation*}
Thus, estimating the covariance function can also be viewed as estimating the mean function of $(X_{i}(s) - \mu(s))(X_{i}(t) - \mu(t))$.

To obtain the qualified summary, one can also apply quadratic regularization. For the ICLP covariance function $K$, denote the tensor product space as
\begin{equation*}
    \mcH_{K^{\eta}\otimes K^{\eta}}: = \mcH_{K^{\eta}}  \otimes  \mcH_{K^{\eta}}, 
\end{equation*}
and RKHS associated with reproducing kernel
\begin{equation*}
    K^{\eta}\otimes K^{\eta}((s_{1},t_{1}),(s_{2},t_{2})) = K^{\eta}(s_{1},s_{2})K^{\eta}(t_{1},t_{2}).
\end{equation*}
With slight abuse of notation, we let $\otimes$ also denote the tensor product of elements in $\mbH$, and then the ICLP with quadratic regularization can be expressed as 
\begin{equation*}
    \hat{C} = \underset{C \in \mcH_{K^{\eta}\otimes K^{\eta}} }{\operatorname{argmin}} \left\{ \frac{1}{n}\sum_{i=1}^{n} \left\| (X_{i}-\bar{X})\otimes (X_{i}-\bar{X})  - C \right\|_{\mbH\otimes\mbH}^2 + \psi \|C\|_{\mcH_{K^{\eta}\otimes K^{\eta}}}^2 \right\}
\end{equation*}
which leads to
\begin{equation*}
    \hat{C}  = \left( K^{\eta}\otimes K^{\eta} + \psi \mathbf{I} \right)^{-1} K^{\eta}\otimes K^{\eta}(\bar{C}).
\end{equation*}
Further assuming $\{\lambda_{j}\}_{j\geq 1}$ and $\{\phi_{j}\}_{j\geq 1}$ as the eigenvalues and eigenfunctions of $K$, then
\begin{equation*}
    \hat{C}(s,t)  = \sum_{j,l\geq 1} \frac{\lambda_{l}^{\eta}\lambda_{l}^{\eta}}{\lambda_{l}^{\eta}\lambda_{l}^{\eta} + \psi} \left\langle \bar{C}, \phi_{j}\phi_{l} \right\rangle_{\mbH\otimes\mbH} \phi_{j}(s)\phi_{l}(t).
\end{equation*}
The expression of $\hat{C}$ is analogous to the expression of $\hat{\mu}$ under the quadratic regularization in Section~\ref{proof: GS analysis for mean protection}. Therefore, similar global sensitivity and utility analysis can be applied to $\hat{C}$ as well since privatizing the covariance function is using the same ICLP via tensor basis, i.e., 
\begin{equation*}
    X(s,t) = \sum_{j,l\geq 1} \sqrt{\lambda_{k}\lambda_{l}} Z_{kl} \phi_{k}(s) \phi_{l}(t)
\end{equation*}
where $Z_{k,l}$ are i.i.d. Laplace random variables with mean $0$ and variance $1$.

\subsection{Function-on-Scalar Linear Regression}
We consider the following Function-on-Scalar linear regression, i.e., 
\begin{equation*}
    Y_{i}(t) = X_{i}^{T} \beta(t) + \epsilon_{i}(t),\quad \text{for} \quad i = 1,\cdots,n
\end{equation*}
where $Y_{i}(t)$ and $e_{i}(t)$ are functional response and error that lies in Hilbert space $\mbH$, covariates $X_{i}\in \mbR^{p}$ and coefficient function $\beta(t) \in \mbH^{\otimes p}$ when $\mbH^{\otimes p}$ denotes $p$-fold Cartesian product of $\mbH$. Estimating and privatizing the coefficient functions $\beta(t)$ is of primary interest. 

One can obtain the estimation via the classical ordinary least square (OLS) estimator, i.e., 
\begin{equation*}
    \hat{\beta} = \underset{\beta\in \mbH^{\otimes p}}{\operatorname{argmin}} \frac{1}{n} \sum_{i=1}^{n} \left\| Y_{i} - X_{i}^{T} \beta \right\|_{\mbH}^2.
\end{equation*}
The OLS estimator is then
\begin{equation*}
    \hat{\beta}(t) = \left( \frac{1}{n} \sum_{i=1}^{n} X_{i} X_{i}^{T} \right)^{-1} \left( \frac{1}{n} \sum_{i=1}^{n} Y_{i}(t) X_{i} \right).
\end{equation*}
Noticing the functional components involved in $\hat{\beta}(t)$ are $Y_{i}(t)$, thus estimating the coefficient function can also be viewed as estimating the mean function of $Y_{i}(t)X_{i}$ and the applied the inverse matrix with scalar elements.

In classical simple linear regression, where both response and covariate are scalars, i.e.
\begin{equation*}
    y_{i} = x_{i}\beta + \epsilon_{i},\quad \text{for} \quad i=1,\cdots,n.
\end{equation*}
Let $\mathbf{x} := (x_{1},\cdots,x_{n})$ and $\mathbf{y} := (y_{1},\cdots,y_{n})$
To achieve $\epsilon$-DP for the OLS estimator, one typically privatizes the empirical variance of $\mathbf{x}$ and the empirical covariance between $\mathbf{x}$ and $\mathbf{y}$, rather than privatizing the $\hat{\beta}$ directly, see \citet{alabi2020differentially}. Specifically, let $\bar{x} = \frac{1}{n}\sum_{i=1}^{n}x_{i}$, $\bar{y} = \frac{1}{n}\sum_{i=1}^{n}y_{i}$, $n \text{cov}(\mathbf{x},\mathbf{y}) = \langle \mathbf{x} - \bar{x}\mathbf{1}, \mathbf{y} - \bar{y}\mathbf{1}\rangle_{\ell^2}$, and $n\text{var}(\mathbf{x}) = \langle \mathbf{x} - \bar{x}\mathbf{1}, \mathbf{x} - \bar{x}\mathbf{1}\rangle_{\ell^2}$. Assuming the sensitivity of $n \text{cov}(\mathbf{x},\mathbf{y})$ and $n\text{var}(\mathbf{x})$ are both $1$ without losing generality. Then, the private OLS estimator that achieves $\epsilon$-DP is
\begin{equation*}
    \tilde{\beta} = \frac{n \text{cov}(\mathbf{x},\mathbf{y}) + Z_{1}}{n\text{var}(\mathbf{x}) + Z_{2}}
\end{equation*}
where $Z_{1}$ and $Z_{2}$ are independent random variables generated from $Lap(0,\frac{1}{\gamma\epsilon})$ and $Lap(0,\frac{1}{(1-\gamma)\epsilon})$ respectively with $\gamma \in (0,1)$.

Turning back to the Function-on-Scalar regression case, one can privatize the statistic $T_{1}:=\frac{1}{n} \sum_{i=1}^{n} X_{i} X_{i}^{T}$ and $T_{2}:=\frac{1}{n} \sum_{i=1}^{n} Y_{i}(t) X_{i}$ separately by splitting the privacy budget $\epsilon$ in to $\gamma \epsilon$ (for $T_{1}$) and $(1-\gamma)\epsilon$ (for $T_{2}$). While $T_{1}$ is a $p\times p$ matrix with scalar elements, one can privatize it with a classical multivariate privacy tool with budget $\gamma\epsilon$. For the statistic $T_{2}$, it is the mean function of $Y_{i}(t)X_{i}$ and thus, we can directly apply the mean function protection we develop in Section~\ref{sec: mean function protection}.

\section{Additional Results for Mean Protection}\label{apd: additional experiments for mean function}
This section presents additional experimental results that are conducted in different settings for mean protection. First, we consider generating the error function from a basis expansion instead of a stochastic process. Second, we set the covariance function of the ICLP as the Mat\'ern kernel with $\alpha = 2.5$, resulting in $\beta = 3$. The Mat\'ern kernel has been widely used to control the eigenvalue decay rate in kernel-based methods; see \citet{wang2022gaussian,lin2024on,lin2024smoothness}

\subsection{Results for Different Error Function}
\begin{figure}[!htbp]
    \centering
    \begin{subfigure}{1\textwidth}
      \centering
      \includegraphics[width=\linewidth]{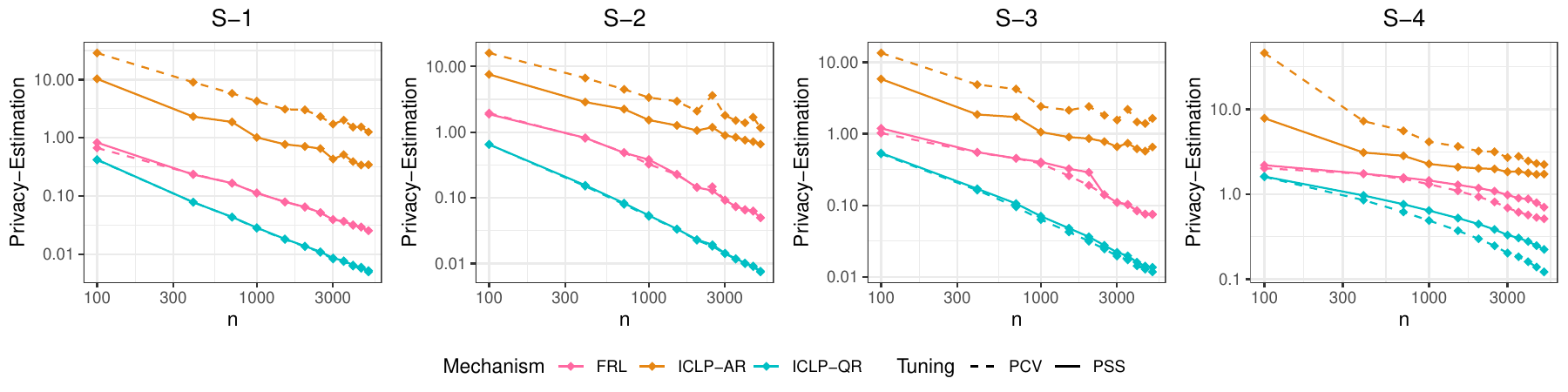} 
      \caption{Privacy-estimation error for PSS and PCV.}
    \end{subfigure}
    \newline
    \begin{subfigure}{1\textwidth}
      \centering
      \includegraphics[width=\linewidth]{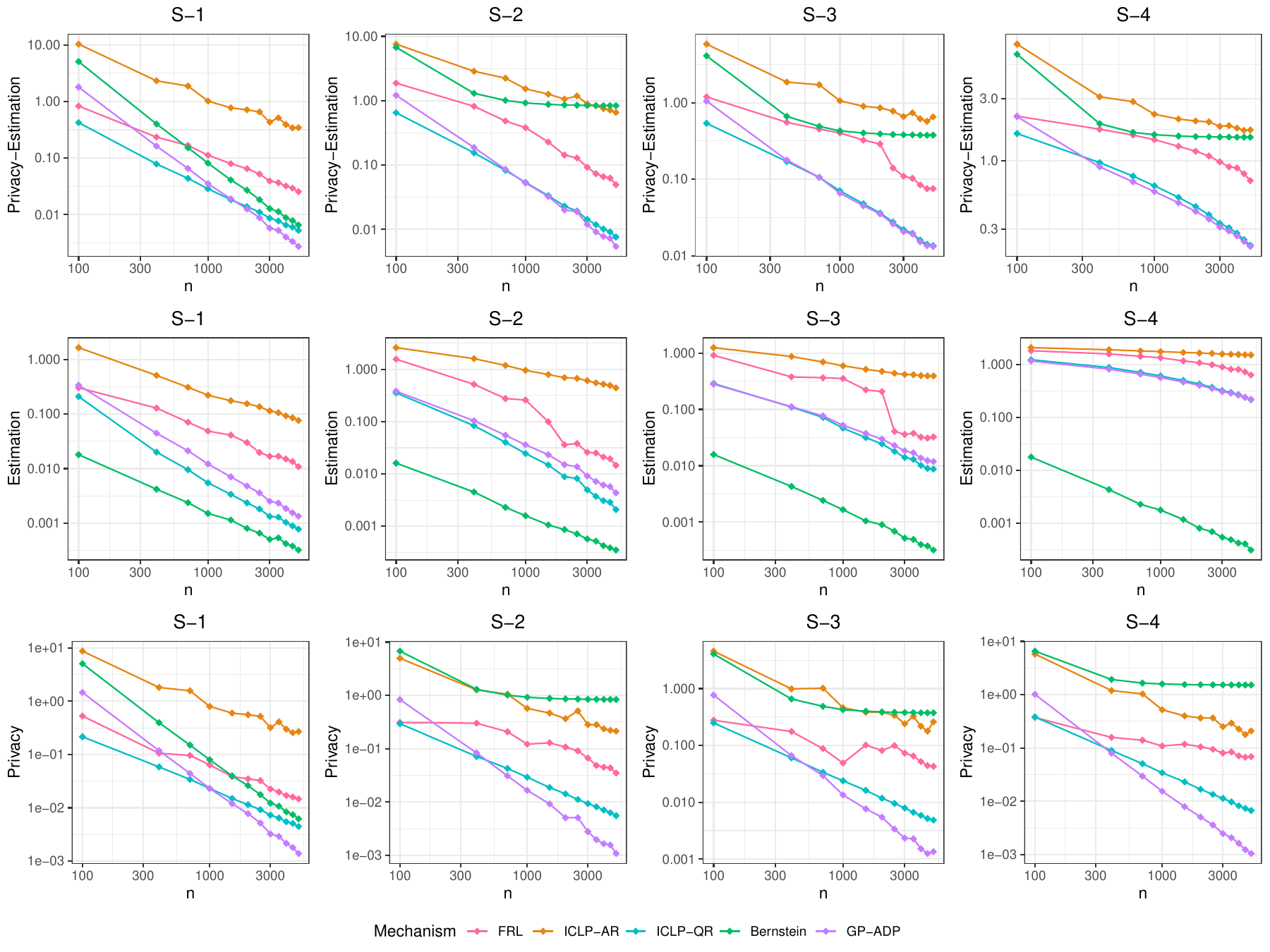}  
      \caption{Privacy-estimation, estimation, and privacy errors with PSS.}
    \end{subfigure}
    \caption{Error decay curves for different mechanisms, sample sizes, and true mean functions. The ICLP covariance function is the Mat\'ern Kernel $C_{\frac{3}{2}}$, and the error function $e_{i}$ is generated via basis expansion with coefficients randomly drawn from a $t$ distribution.}
    \label{fig: error via basis expansion and t distribution}
\end{figure}

We generate the functional sample curves as 
\begin{equation*}
    X_{i}(t) = \mu_{0}(t) + e_{i} \quad \text{where} \quad e_{i}(t) = \sum_{j=1}^{100}U_{ij} \phi_{j}(t)
\end{equation*}
where $U_{ij}$ are i.i.d. random variables from a $t$ distribution with degree of freedom $5$. We conduct the same experiments and report the results in Figure~\ref{fig: error via basis expansion and t distribution}. One interesting observation is that when $e_{i}$ are generated from a basis expansion with heavy-tailed coefficients, the PSS and PCV are more aligned with each other in S-2 and S-3.

\subsection{Results for Different ICLP Covariance}
We set the ICLP covariance function as the Mat\'ern kernel with $\alpha = 2.5$, resulting in $\lambda_j \asymp j^{-6}$. We repeat the comparison between PSS and PCV and the experiments that compare different mechanisms under different $n$. 

\begin{figure}[!htbp]
    \centering
    \begin{subfigure}{1\textwidth}
      \centering
      \includegraphics[width=\linewidth]{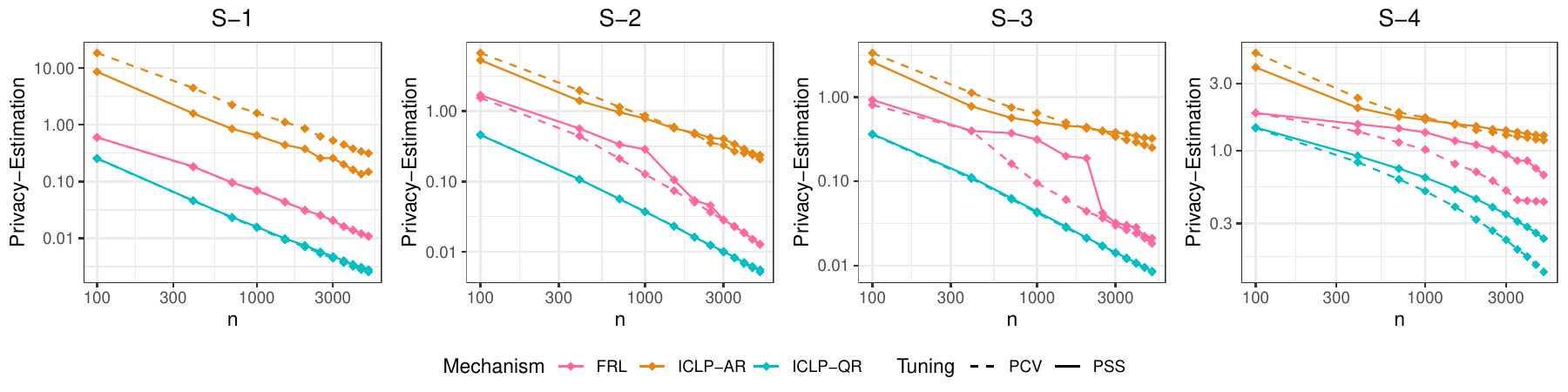} 
      \caption{Privacy-estimation error for PSS and PCV.}
    \end{subfigure}
      \centering
      \begin{subfigure}{1\textwidth}
      \includegraphics[width=\linewidth]{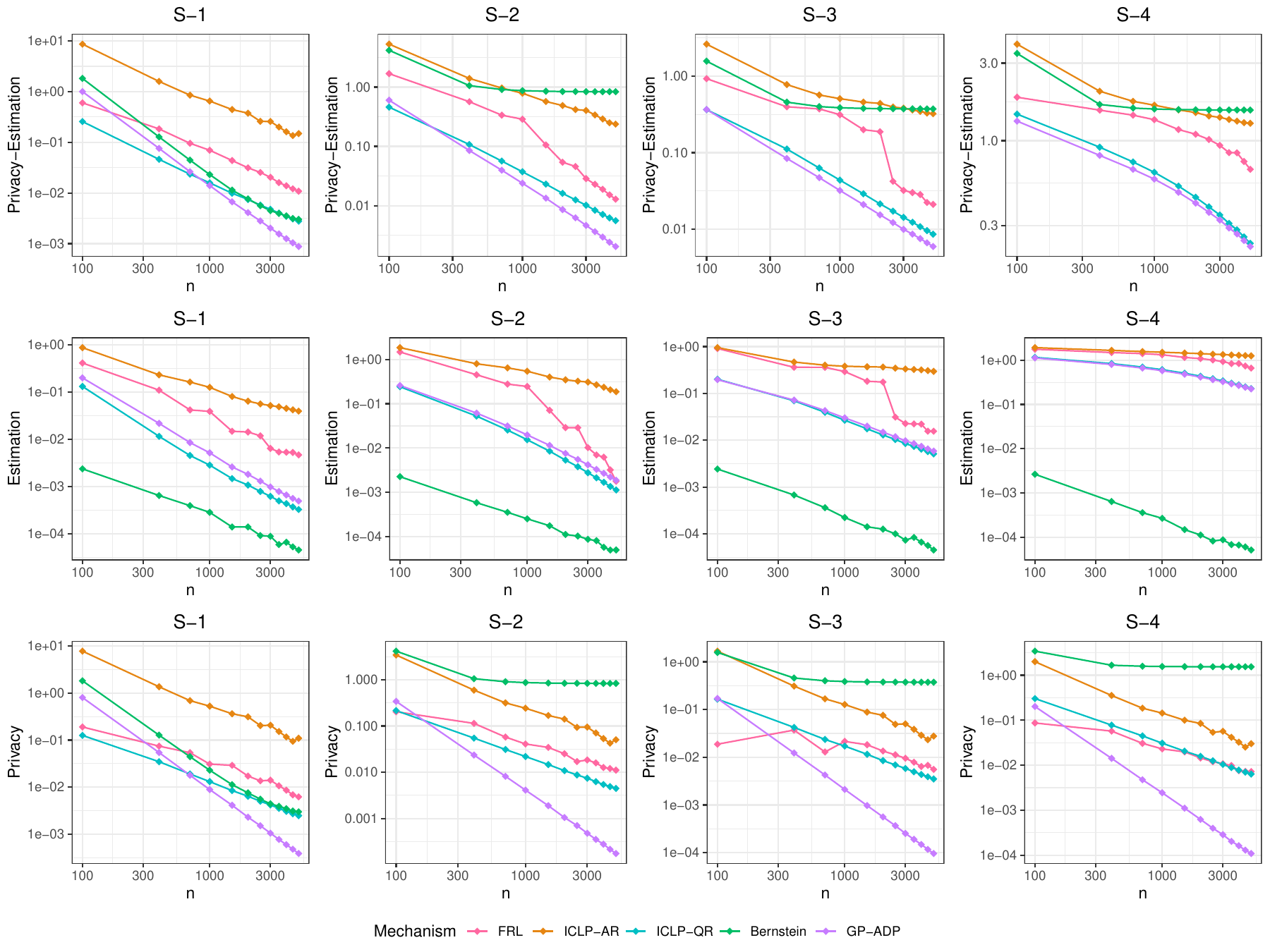}  
      \caption{Privacy-estimation, estimation, and privacy errors with PSS.}
      \end{subfigure}
    \caption{Error decay curves for different mechanisms, sample sizes, and true mean functions. The ICLP covariance function is the Mat\'ern Kernel $C_{\frac{5}{2}}$, and the error function $e_{i}$ is generated from the Gaussian process with RBF kernels.}
    \label{fig: error with gaussian kernel for nu52}
\end{figure}

\begin{figure}[!htbp]
    \centering
    \begin{subfigure}{1\textwidth}
      \centering
      \includegraphics[width=\linewidth]{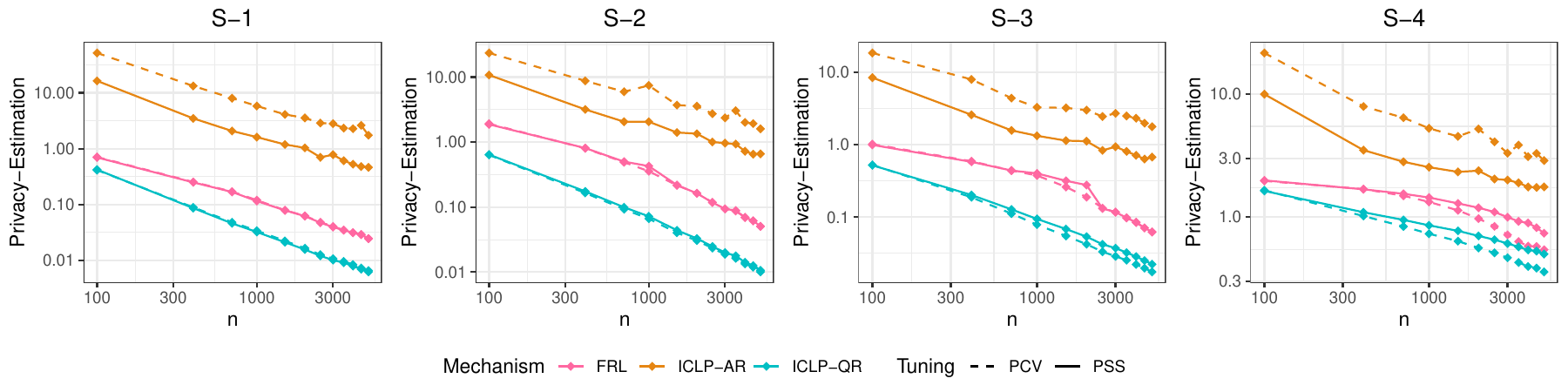} 
      \caption{Privacy-estimation error for PSS and PCV.}
    \end{subfigure}
    \begin{subfigure}{1\textwidth}
      \centering
      \includegraphics[width=\linewidth]{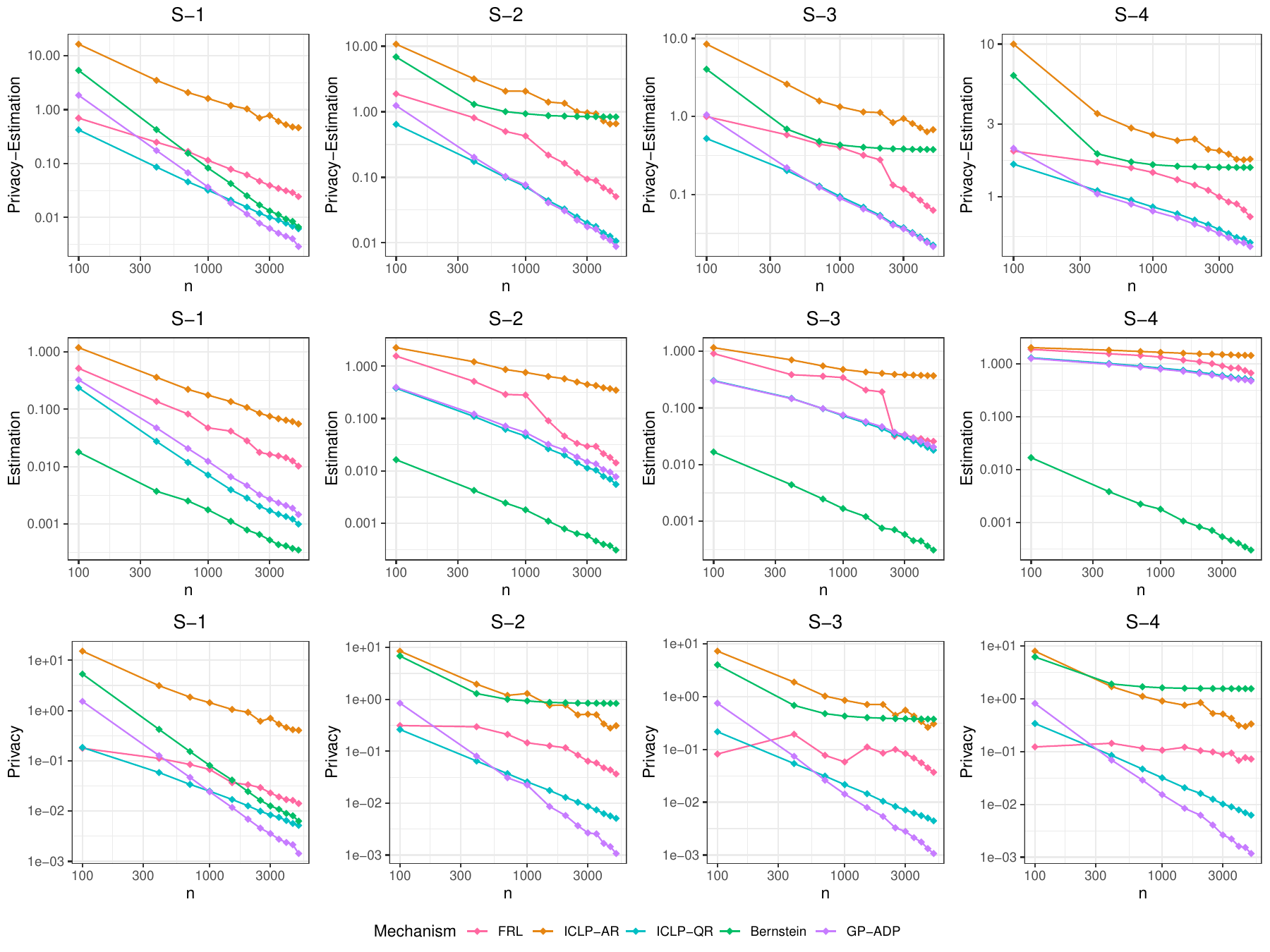}  
      \caption{Privacy-estimation, estimation, and privacy errors with PSS.}
      \end{subfigure}
    \caption{Error decay curves for different mechanisms, sample sizes, and true mean functions. The ICLP covariance function is the Mat\'ern Kernel $C_{\frac{5}{2}}$, and the error function $e_{i}$ is generated via basis expansion with coefficients randomly drawn from a $t$ distribution.}
    \label{fig: error via basis expansion and t distribution for nu52}
\end{figure}

The results are reported in Figure~\ref{fig: error with gaussian kernel for nu52} for the error function as a Gaussian stochastic process and Figure~\ref{fig: error via basis expansion and t distribution for nu52} for the error function generated from a basis expansion. It can be observed from the figure that the results based on $C_{\frac{5}{2}}$ are almost the same as those based on $C_{\frac{3}{2}}$.

\newpage
\section{Visualization of the Human Mortality Application}\label{apd: Visualization of Age-at-Death KDE}
\begin{figure}[ht]
    \centering
    \includegraphics[width = 0.85\textwidth]{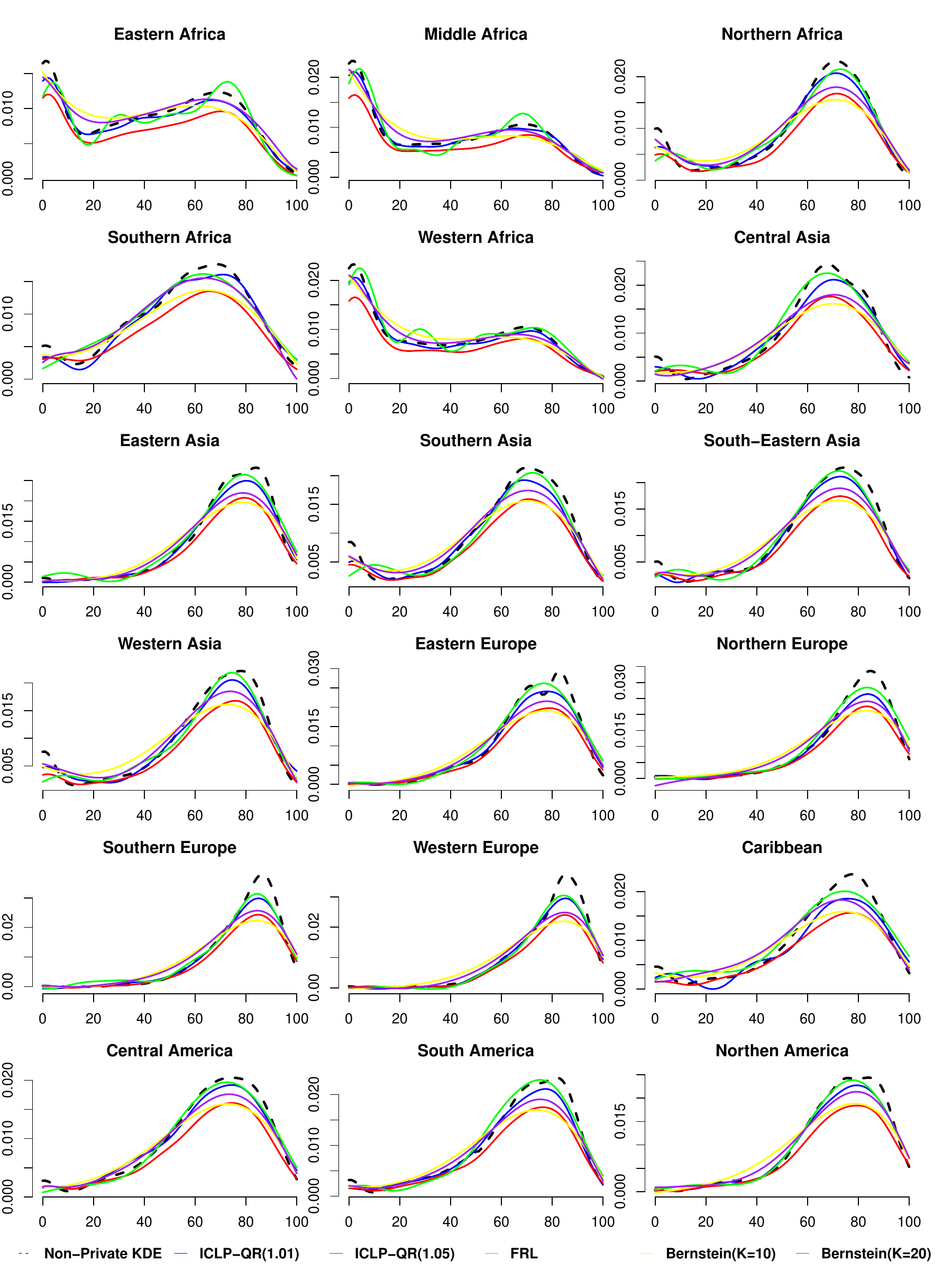}
    \caption{Non-private and private kernel density estimates of age-at-death density in different regions under different mechanisms with $\epsilon = 1$.}
    \label{fig: KDE visualization}
\end{figure}

In Section~\ref{subsec: age-at-death KDE}, we only reported the expected $L^{2}$-distance of private KDEs to their non-private counterparts. Here, we present the visualization of the comparison between non-private KDEs and private KDEs for different mechanisms in Figure~\ref{fig: KDE visualization}.

\clearpage
\vskip 0.2in
\bibliography{ref}

\end{document}